\documentclass[10pt,journal,compsoc]{IEEEtran}
\usepackage{graphicx}
\usepackage{comment}
\usepackage{amsmath,amssymb} 
\usepackage{multirow}
\usepackage{floatrow}
\usepackage{color}
\usepackage{float}
\usepackage{caption}
\usepackage{color}
\usepackage{tablefootnote}
\usepackage{paralist}
\usepackage{hyperref}
\ifCLASSOPTIONcompsoc
  \usepackage[nocompress]{cite}
\else
  \usepackage{cite}
\fi
\ifCLASSINFOpdf
\else
\fi

\begin{document}
%
\title{VPN++: Rethinking Video-Pose embeddings for understanding Activities of Daily Living}
%
%
%
%

\author{\normalsize Srijan Das,
        Rui Dai, 
        Di Yang,
        Francois Bremond
\IEEEcompsocitemizethanks{\IEEEcompsocthanksitem S. Das is with Stony Brook University, 100 Nicolls Rd, Stony Brook, NY 11794, USA.\protect\\
E-mail: srijan.das@stonybrook.edu}
\IEEEcompsocitemizethanks{\IEEEcompsocthanksitem R. Dai, D. Yang, and F. Bremond are with the Inria and Universite Cote d'Azur, 2004 Route des Lucioles, 06902 Valbonne, France.\protect

E-mail: \{rui.dai, di.yang, francois.bremond\}@inria.fr}}
\IEEEtitleabstractindextext{%
\begin{abstract}
Many attempts have been made towards combining RGB and 3D poses for the recognition of Activities of Daily Living (ADL).  ADL may look very similar and often necessitate to model fine-grained details to distinguish them. Because the recent 3D ConvNets are too rigid to capture the subtle visual patterns across an action, this research direction is dominated by methods combining RGB and 3D Poses. 
But the cost of computing 3D poses from RGB stream is high in the absence of appropriate sensors. This limits the usage of aforementioned approaches in real-world applications requiring low latency. Then, how to best take advantage of 3D Poses for recognizing ADL?

To this end, we propose an extension of a pose driven attention mechanism: Video-Pose Network (VPN), exploring two distinct directions. One is to transfer the Pose knowledge into RGB through a feature-level distillation and the other towards mimicking pose driven attention through an attention-level distillation. Finally, these two approaches are integrated into a single model, we call \textbf{VPN++}.  We show that VPN++ is not only effective but also provides a high speed up and high resilience to noisy Poses. VPN++, with or without 3D Poses, outperforms the representative baselines on 4 public datasets. Code is available at \href{https://github.com/srijandas07/vpnplusplus}{https://github.com/srijandas07/vpnplusplus}.
\end{abstract}


}

\newcommand{\datasetfullname}{Toyota Smarthome Untrimmed}
\newcommand{\datasetinitialname}{TSU}
\newcommand{\methodfullname}{Attention Guided Net}
\newcommand{\methodinitialname}{AGNet}

\definecolor{srijan}{rgb}{0,0,1}
\newcommand{\commsrijan}[1]{\textcolor{srijan}{{#1}}}
\maketitle

\IEEEdisplaynontitleabstractindextext

%
\IEEEpeerreviewmaketitle

\IEEEraisesectionheading{\section{Introduction}\label{sec:introduction}}

\IEEEPARstart{L}{earning} representations for human actions, taking into account only the RGB modality is not sufficient. As a consequence, a large corpus of research studies has been focusing on multi-modal action recognition. 
The most popular and effective method is the two-stream approach~\cite{twostream, twostreamfusion, i3d} where one stream models appearance by taking RGB frames and the other stream models short-term motion by taking optical flow frames. 
However, this method is effective on videos obtained from web~\cite{ucf, kinetics, kuehne2011hmdb} where the human actions have prominent motion patterns. But what about Activities of Daily Living (ADL) where actions have subtle motion and often pertain to have similar spatio-temporal patterns?
\begin{figure}
\begin{center}
   \includegraphics[width=1\linewidth]{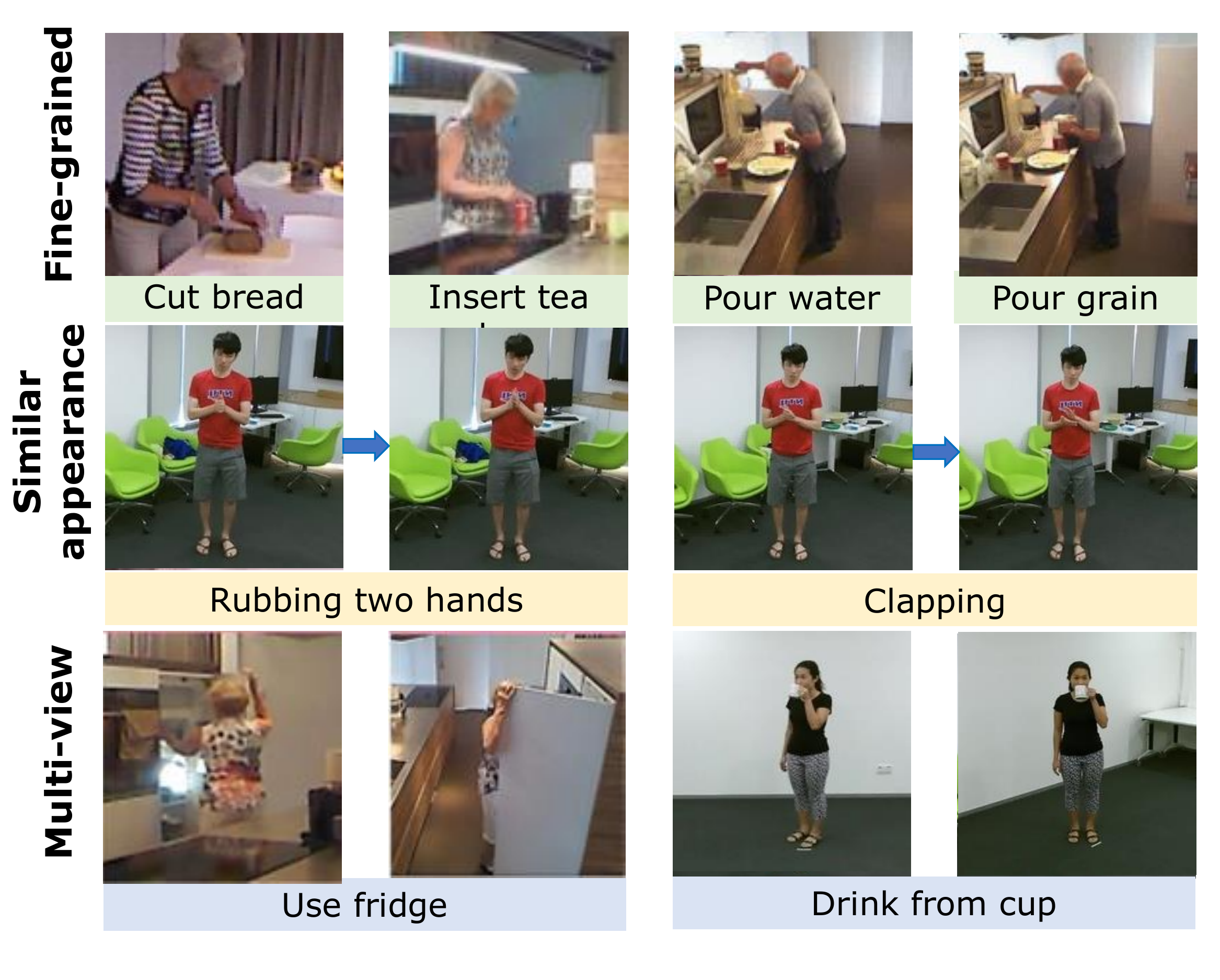}
\end{center}
   \caption{Illustration of the challenges in Activities of Daily Living: fine-grained actions (top),  actions with similar visual pattern (middle) and actions viewed from different cameras (below).}
\label{samples}
\end{figure}

\begin{figure}
\begin{center}
   \includegraphics[width=1\linewidth]{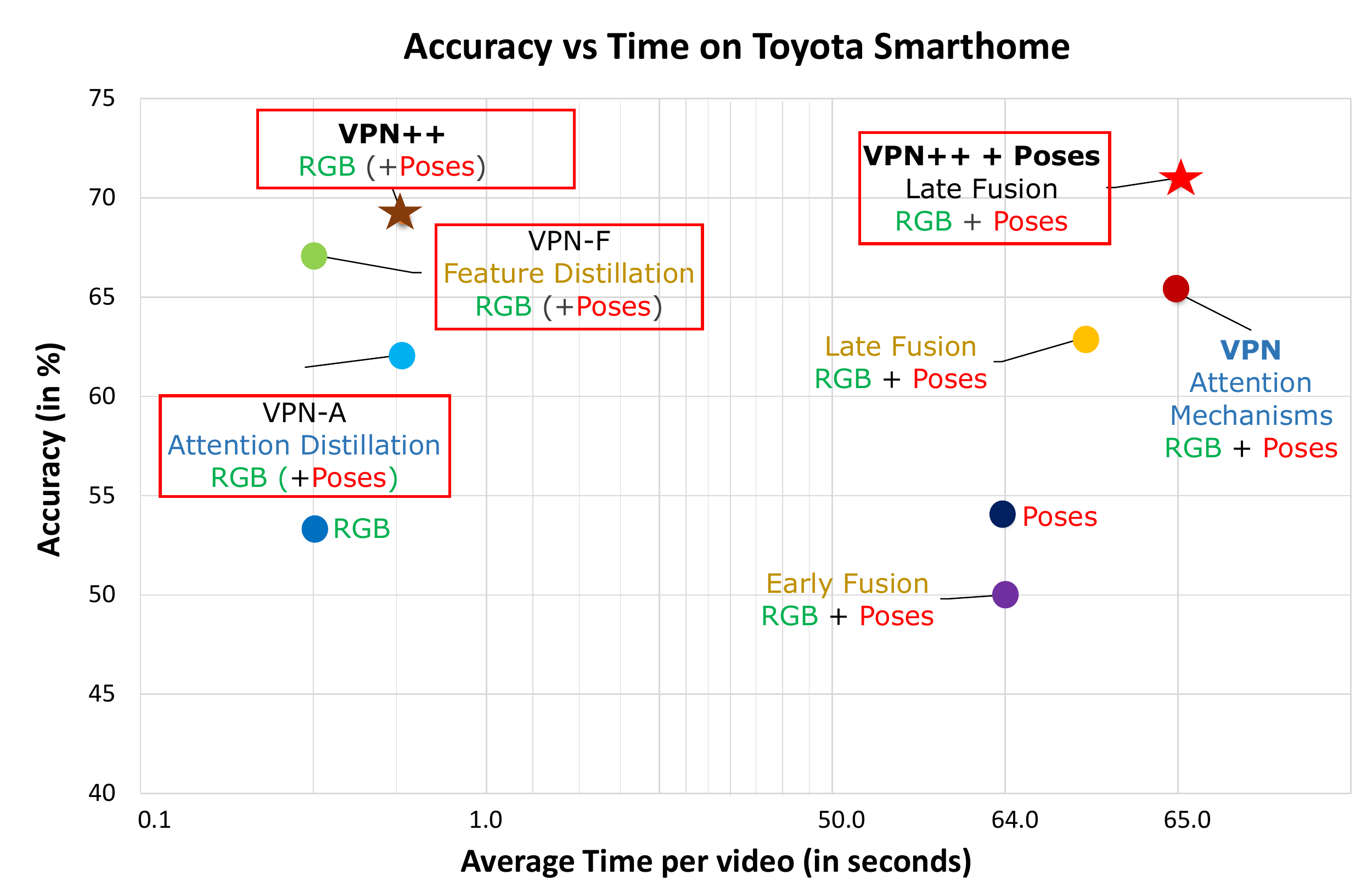}
\end{center}
   \caption{Accuracy vs Time plot on Toyota Smarthome dataset for RGB and Pose modalities. 3D Poses are estimated using LCRNet++~\cite{lcrnet_new} followed by Videopose3D~\cite{videopose3d}. Early fusion indicates concatenation of features at the last layer before prediction whereas Late fusion indicates averaging the prediction from both modalities.
   Our proposed models (marked with bounding box): \textbf{VPN-F}, \textbf{VPN-A} and \textbf{VPN++} mimicking Pose stream, outperforms all other RGB and Pose combining strategies, while being significantly faster. Late fusion of the distilled models with Pose stream further boosts the classification accuracy, but at the price of the model efficiency. Note that the model with input modalities denoted by \textcolor{green}{RGB} \textcolor{red}{(+Poses)} have been trained with RGB and Poses but do not require Poses at inference time.}
\label{time_analysis_intro}
\end{figure}

Activities of Daily Living (ADL) may look simple but their recognition is often more challenging than activities present in sport, movie or Youtube videos. 
ADL often have very low inter-class variance making the task of discriminating them from one another very challenging. The challenges characterizing ADL are illustrated in fig~\ref{samples}: (i)  short and subtle actions like \textit{pouring water} and \textit{pouring grain} while \textit{making coffee} ;
(ii) actions exhibiting similar visual patterns while differing in motion patterns like \textit{rubbing hands} and \textit{clapping}; and finally, (iii) actions observed from different camera views.
In the recent literature, the main focus is the recognition of actions from internet videos~\cite{i3d,nonlocal,slow_fast,TSN,video_transformer_network} and very few studies have attempted to recognize ADL in indoor scenarios~\cite{timeception,glimpse,STA_iccv}. For instance, state-of-the-art 3D convolutional networks like I3D~\cite{i3d} pre-trained on huge video datasets~\cite{kinetics,ucf,kuehne2011hmdb} have successfully boosted the recognition of actions from internet videos. But, these networks with similar spatio-temporal kernels applied across the whole space-time volume cannot address the complex challenges exhibited by ADL. Attention mechanisms have thus been proposed on top of these 3D convolutional networks to guide them along the regions of interest of the targeted actions~\cite{nonlocal,timeception,video_transformer_network}.

Towards another approach, recent studies~\cite{NTU_RGB+D, ntu120, STA_iccv} have shown that human 3D poses provide a strong clue for understanding human-centric patterns in videos. Of course the use of 3D poses for human action analysis depends on (i) the availability of good quality 3D poses and (ii) architectures processing them. Thanks to algorithms like LCRNet++~\cite{lcrnet_new} and VideoPose3D~\cite{videopose3d}, high quality 3D poses can be obtained from RGB without the requirement of depth sensors. Similarly, the advancement of Graph based CNN architectures~\cite{stgcn2018aaai, directed_graph, 2sagcn2019cvpr, msaagcn} that take into account the human joint configurations have greatly impacted the skeleton based action recognition. Since skeleton based action recognition does not leverage the appearance information in videos, combining 3D poses and RGB is the need of the hour as studied in~\cite{rgb+pose_1, rgb+pose_2, rgb+pose_3, nas_multi, glimpse, Baradel_BMVC, STA_hands, spatial-i3d}.

The most common strategy for combining RGB stream and 3D poses includes (i) feature or score level fusion~\cite{rgb+pose_1, rgb+pose_2, rgb+pose_3, nas_multi}. As these modalities are heterogeneous, they must be processed by different kinds of network to show their effectiveness. This limits their performance in simple multi-modal fusion strategy~\cite{rgb+pose_1,rgb+pose_2,Luo_2018_ECCV}. Therefore, another approach adopted in recent days includes (ii) pose driven attention mechanisms~\cite{glimpse, Baradel_BMVC, STA_hands, spatial-i3d}. However, these methods have improved the action recognition performance but they do not take into account the alignment of the RGB cues and the corresponding 3D poses. Therefore, we proposed a spatial embedding to project the visual features and the 3D poses in the same referential in~\cite{VPN}.

Further, this embedding is accompanied by an attention network to recognize a large variety of human actions. Thus, VPN consists of a spatial embedding and an attention network.
It exhibits the following properties through its modules:
(i) a spatial embedding learns an accurate video-pose embedding to enforce the relationships between the visual content and 3D poses, (ii) an attention network learns the attention weights with a tight spatio-temporal coupling for  better modulating the RGB feature map, (iii) the attention network takes the spatial layout of the human body into account by processing the 3D poses through Graph Convolutional Networks (GCNs).

VPN to some extent overcomes the challenge of combining two modalities that are not only semantically different but also processed through heterogeneous networks. To go beyond these approaches, we study novel manners to combine efficiently RGB and 3D Poses.
In particular, we aim at relaxing the need of high quality 3D poses, which are not always available.
In Figure~\ref{time_analysis_intro}, we provide a plot of action classification accuracy vs average inference time on Toyota Smarthome~\cite{STA_iccv} dataset. From the plot, we observe that feature level fusion (Early Fusion) performs worse since such fusion mechanisms are often prone to over-fitting~\cite{mutli-modal_harder} owing to an increase in the number of parameters of the network. 
Besides, Pose driven attention mechanism~\cite{VPN} yields high classification accuracy compared to RGB~\cite{i3d} and Poses~\cite{2sagcn2019cvpr} individually or their score level fusion (Late Fusion). But these RGB+Poses based methods are significantly slower than the RGB ones.

To this end, we explore the concept of knowledge distillation to infuse pose stream into RGB stream. Towards this objective, we propose two levels of distillation - one taking an approach of feature level fusion and the other one benefiting from attention mechanism.
First, we aim at transferring feature-level knowledge from Pose to RGB stream to learn discriminative representation for recognizing actions, we call this feature-level distillation model \textbf{VPN-F}. 
To learn VPN-F, we use contrastive learning for distilling the knowledge from Pose stream to RGB.
Besides avoiding the computation of poses at inference time, VPN-F learns to maximize the salient information from both streams towards action recognition.
Second, we mimic pose driven attention network as in VPN through RGB stream. This is performed by adding a self-attention block in the RGB stream that hallucinates attention weights learned through 3D poses for the task of action recognition. We call this attention-level distillation model \textbf{VPN-A}. 
As an end result, VPN-A learns to provide pose driven attention weights which not only improve the action classification accuracy but also eliminate the requirement of poses at inference time.
Finally, we integrate both levels of distillation into a single model called \textbf{VPN++}.
Our experiments confirm that VPN++ is 160 times faster than the state-of-the-art methods without compromising effectiveness in real-world scenarios as illustrated in fig.~\ref{time_analysis_intro}.
We also show that VPN++ via distillation when combined with 3D Poses, if available, outperforms the state-of-the-art results on 4 public datasets. Thus, to sum up, by infusing Poses into RGB using distillation, we provide a choice of highly effective models to the community that can be leveraged based on their needs like low latency, low sensitivity towards noisy Poses, or none. 

\section{Related Work}
  Significant improvement has been made in the action recognition domain after the advancement of 3D CNN~\cite{C3D}. Carreira and Zisserman~\cite{i3d} proposed a 3D CNN based fully convolutional network namely I3D which is pre-trained on huge datasets like Kinetics~\cite{kinetics} to capture discriminative spatio-temporal patterns within an action. With the success of I3D, holistic methods like Pseudo 3D CNN~\cite{pseudo3d}, Separable 3D CNN~\cite{can_spatio-temporal}, slow-fast network~\cite{slow_fast}, channel-separated CNN~\cite{CSN}, and X3D~\cite{x3d}  have been fabricated for generic video datasets like Kinetics~\cite{kinetics}, UCF-101~\cite{ucf} and HMDB~\cite{kuehne2011hmdb}.
 But these networks with similar kernels applied across the whole space-time volume of a video, are too rigid to capture salient features for subtle patterns in ADL. Recently several attention mechanisms have been proposed on top of the aforementioned 3D ConvNets to extract salient spatio-temporal patterns. For instance, Wang et al.~\cite{nonlocal} have proposed a non-local module on top of I3D which computes the attention of each pixel as a weighted sum of the features of all pixels in the space-time volume. But this module relies too much on the appearance of the actions, i.e., pixel position within the space-time volume. As a consequence, this module though effective for the classification of actions in internet videos, fails to disambiguate ADL with similar motion and fails to address view-invariant challenges.
 
 On the other hand, temporal evolution of 3D poses has been leveraged through sequential networks like LSTM and GRU for skeleton based action recognition~\cite{gemetricfeaturesWACV2017,st-lstm,valstm}. Taking a step ahead, LSTMs have also been used for spatial and temporal attention mechanisms to focus on the salient human joints and key temporal frames~\cite{sta_lstm}. Another framework represents 3D poses as pseudo images to leverage the successful image classification CNNs for action classification~\cite{skel_CNN_1,skel_CNN_2}.
 Moreover, skeleton based action recognition has made significant improvements with the advancement of Graph Convolutional Networks (GCNs)~\cite{stgcn2018aaai, directed_graph, 2sagcn2019cvpr, msaagcn}. The key idea is to feed a graph representation of a skeleton frame in these networks which are optimized for the task of action classification. These graph based methods make use of the spatial topology of the human body joints and thus are more effective than recurrent networks~\cite{NTU_RGB+D, st-lstm}. However, the skeleton based action recognition lacks in encoding the appearance information which is critical for ADL, such as in human-object interactions.

 \noindent \textbf{Combining modalities}: Combining the advantages of privileged modalities in order to make use of their complementary discriminative power has been exploited widely in action recognition domain. Two-stream architectures~\cite{twostream, twostreamfusion, i3d} that learn separate features from optical flow and RGB modalities, outperform single modality approaches. Towards this direction, Ryoo et al.~\cite{Ryoo2020AssembleNet:, assemblenetplusplus} have proposed a Neural Search Architecture (NAS) to combine both RGB and Optical flow streams. In contrast to these methods, two complementary strategies are adopted to combine RGB and pose modalities. One is fusion of both modalities in feature space~\cite{rgb+pose_1, rgb+pose_2, rgb+pose_3, nas_multi}. However, these modalities are heterogeneous and must be processed by different kinds of network to show their effectiveness.  Combining these heterogeneous features from different modalities through feature/score fusion introduce noise resulting in a downgraded action recognition performance~\cite{srijan_MM}.
 The second is pose driven attention mechanisms to guide the RGB cues for action recognition as in~\cite{Baradel_BMVC, STA_hands, spatial-i3d, STA_iccv}. In~\cite{STA_hands,Baradel_BMVC,glimpse}, the pose driven attention networks implemented through LSTMs, focus on the salient image features and the key frames. Then, with the success of 3D CNNs, 3D poses have been exploited to compute the attention weights of a spatio-temporal feature map.
Das et al.~\cite{spatial-i3d} have proposed a spatial attention mechanism on top of 3D ConvNets to weight the pertinent human body parts relevant for an action. Then, authors in~\cite{STA_iccv} have proposed a more general spatial and temporal attention mechanism in a dissociated manner. 
But these methods have the following drawbacks: 
(i) there is no accurate correspondence between the 3D poses and the RGB cues in the process of computing the attention weights~\cite{STA_hands,Baradel_BMVC,glimpse,spatial-i3d,STA_iccv};
(ii) the attention sub-networks~\cite{STA_hands,Baradel_BMVC,glimpse,spatial-i3d,STA_iccv} neglect the topology of the human body while computing the attention weights;
(iii) the attention weights in~\cite{spatial-i3d,STA_iccv} provide identical spatial attention along the video. As a result, action pairs with similar appearance like \textit{jumping} and \textit{hopping} are mis-classified. Therefore in~\cite{VPN}, we proposed a new spatial embedding to enforce the correspondences between RGB and 3D poses which has been missing in the state-of-the-art methods. The embedding is built upon an end-to-end learnable attention network. The attention network considers the human topology to better activate the relevant body joints for computing the attention weights. To the best of our knowledge, none of the previous action recognition methods have combined human topology with RGB cues. In addition, the proposed attention network couples the spatial and temporal attention weights in order to provide spatial attention weights varying along time.
 
However, all the above approaches including VPN rely on the availability of 3D Poses, which not only escalates the model inference time but also increases their sensitivity towards Pose quality. Therefore, we use the concept of distillation that not only learns discriminative video-pose representations for understanding actions but also relaxes the demand for Poses at inference time.
Consequently, we adopt both strategies to enforce RGB stream to (i) mimic pose stream features, and (ii) emulate pose-driven attention mechanism. 
 
 \noindent \textbf{Distillation}: Many approaches have exploited the concept of distillation for cross-modal knowledge transfer~\cite{soundnet, hoffman, hoffman_adaptation, garcia_cross_modal_distillation, Luo_2018_ECCV, mars, evolving_losses}. Towards action recognition, Garcia et al.~\cite{garcia_cross_modal_distillation} proposed a distillation framework consisting of teacher-student networks that hallucinates depth features from RGB features. This distillation is performed via logits as well as by matching feature maps of RGB and depth networks. Similarly, distillation approaches dynamically leveraging complementary information across several modalities have been proposed in~\cite{evolving_losses, Luo_2018_ECCV}. 
 Crasto et al.~\cite{mars} proposed MARS to train a RGB stream with standard cross-entropy loss along with mimicking the features learned by an optical flow stream. This mimicking is accomplished by a distillation loss that minimizes the euclidean distance between the learned features across both streams.
 
 Thus, many distillation methods have been studied in the action recognition domain with OF and RGB, but not with RGB and Poses. Infusing Poses into RGB stream through distillation is not straightforward and includes two main challenges:
 (i) 2D RGB images with appearance information and 3D Poses with geometric details are fed to the teacher-student heterogeneous networks, limiting the knowledge transfer between them due to their asymmetric dimensionality; (ii) the teacher network, i.e. the Pose stream is not consistently effective on the entire data distribution. 
 In fact, the Pose stream carries irrelevant features for actions that are discriminated using their appearance information. 
 Therefore, we propose to minimize the distance between the features learned by RGB \& Poses while learning discriminative representation in the RGB feature space. 
 Towards another approach with an effective teacher network, we perform online distillation (via collaborative learning) to transfer pose driven attention knowledge learned from VPN~\cite{VPN} to RGB stream. Distillation methods like~\cite{contrastive_distillation, collaborative} are close to our approaches, however they are specific for image domain applications. In contrast, the extension of VPN: \textbf{VPN++} is dedicated for combining cross-modal information pertaining to video domain applications.
 The feature-level and attention-level distillation mechanisms to infuse Poses into RGB stream through cross-modal knowledge distillation provide a practical model for combining RGB and 3D Poses.

 \section{Video-Pose embedding Models}
In this section, we first detail our previously proposed Video-Pose Network (VPN), followed by an elaborate description of the video-pose embedding models through distillation.
We aim at building a video-pose network \textbf{VPN++} which benefits from two levels of distillation - (i) feature-level, and (ii) attention-level. At training time, the inputs to these models are RGB videos along with their corresponding 3D poses. These 3D poses could be obtained either from Kinect sensors using~\cite{37} or from RGB images using pose estimation algorithms like LCRNet++~\cite{lcrnet_new} and VideoPose3D~\cite{videopose3d}.
The RGB images and the 3D Poses are processed by a video backbone and a pose backbone respectively. In this work, the video backbones are usually 3D CNNs that take as input a stack of human cropped images from a video clip to compute the spatio-temporal representation of the clip. On the other hand, the Pose backbones are  spatio-temporal Graph Convolutional Networks that take a stack of 3D Poses as a graphical input to model actions. 
At inference time, traditional VPN requires both RGB and 3D Poses to predict the actions.
In contrast, VPN++ requires only the RGB videos at inference time to predict the action classes.

\subsection{Background: VPN}
\begin{figure*}
\begin{center}
   \includegraphics[width=1\linewidth]{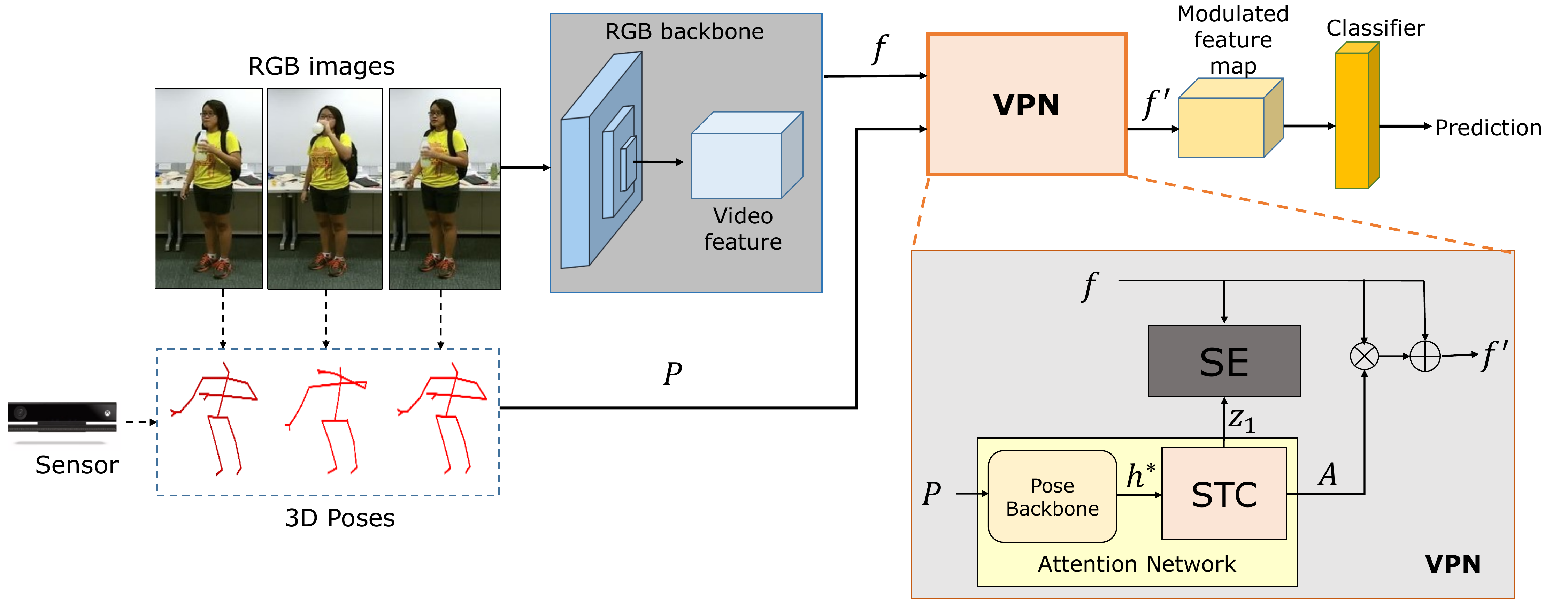}
\end{center}
   \caption{VPN takes as input RGB images with their corresponding 3D poses. The RGB images are processed by a visual backbone which generates a spatio-temporal feature map ($f$). The proposed \textbf{VPN} takes as input the feature map ($f$) and the 3D poses ($P$). VPN consists of two components: an attention network and a spatial embedding (SE). The attention network further consists of a Pose Backbone and STC (spatio-temporal Coupler). VPN computes a modulated feature map $f'$. This modulated feature map $f'$ is then used for classification.}
\label{vpn_fig}
\end{figure*}

VPN can be thought as a layer which can be placed on top of any 3D convolutional backbone. VPN takes as input a 3D feature map ($f \in \mathbb{R}^{c\times t \times m \times n}$) and its corresponding 3D poses ($P$) to perform two functionalities as shown in fig.~\ref{vpn_fig}. First, to provide an accurate alignment of the human joints with the feature map $f$. Second, to compute a modulated feature map ($f'$) which is further classified for action recognition. The modulated feature map ($f'$) is weighted along space and time as per its relevance. VPN exploits the highly informative 3D pose information to transform the visual feature map $f$ and finally, compute the attention weights. This network has two major components: (I) an attention network and (II) a spatial embedding.

\subsubsection{Attention Network}
The attention network consists of a Pose Backbone and a spatio-temporal Coupler (STC).
The input poses along the video are processed in a Pose Backbone as shown in fig~\ref{vpn_fig}.
The pose based inputs of VPN are the 3D human joint coordinates $P \in \mathbb{R}^{3\times J \times t_p}$ stacked along $t_p$ temporal dimension, where $J$ is the number of skeleton joints. The Pose Backbone processes these 3D poses to compute pose features $h^*$ which are used further in the attention network for computing the spatio-temporal attention weights. 

\begin{figure}
\centering
\includegraphics[width=1\linewidth]{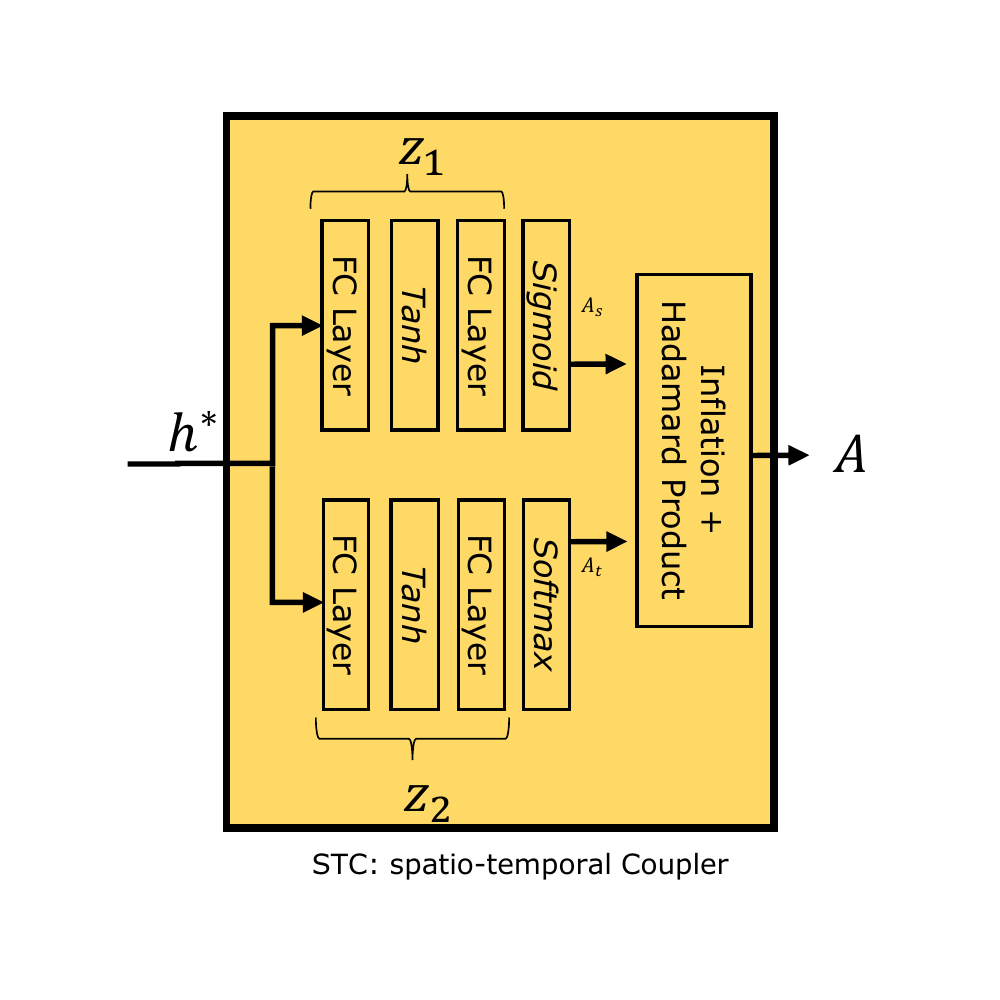} 
\caption{STC: spatio-temporal Coupler to generate spatio-temporal attention weights $A$ from the latent pose based feature $h^*$.}\vspace{-1em}
\label{stc}
\end{figure}

\noindent Next, the attention network in VPN learns the spatio-temporal attention weights $A$ from the output of Pose Backbone in two steps as illustrated in fig.~\ref{stc}. In the first step, $m \times n$ dimensional spatial and $t$ dimensional temporal attention weights are classically trained as in~\cite{sta_lstm} to get the most important body parts and key frames for an action. This learning of spatial and temporal attention weights takes place in two streams ($z_1$ and $z_2$) consisting of dense layers each followed by relevant activations. In the second step, joint spatio-temporal attention weights are computed by performing a Hadamard product on the spatial and temporal attention weights. In order to perform this matrix multiplication, the spatial and temporal attention weights are inflated by duplicating the same attention weights in temporal and spatial dimension respectively.

\noindent This two-step attention learning process enables the attention network to compute spatio-temporal attention weights in which the spatial saliency varies with time. The obtained attention weights are crucial to disambiguate actions with similar appearance as they may have dissimilar motion over time. 
Finally, the spatio-temporal attention weights $A \in \mathbb{R}^{t \times m \times n}$ are linearly multiplied with the input video feature map $f$, followed by a residual connection with the original feature map $f$ to output the modulated feature map $f'$. The residual connection enables the network to retain the properties of the original visual features.

\subsubsection{Spatial Embedding of RGB and Pose}
The objective of the embedding model is to provide tight correspondences between both pose and RGB modalities used in VPN. The state-of-the-art methods~\cite{spatial-i3d,STA_iccv} attempt to provide the attention weights on the RGB feature map using 3D pose information without projecting them into the same 3D referential. The mapping with the pose is only done by cropping the person within the input RGB images.
The spatial attention computed through the 3D joint coordinates does not correspond to the part of the image (no pixel to pixel correspondence), although it is crucial for recognizing fine-grained actions. To correlate both modalities, an embedding technique inspired from image captioning task~\cite{text_video_embedding,joint_modelling_captioning} is used to build an accurate RGB-Pose embedding in order to enable the poses to represent the visual content of the actions.

\noindent Thus, the embedding is performed by propagating a normalized euclidean loss between the visual features and a spatial attention vector ($z_1$ obtained from STC). Both the visual feature and spatial attention vectors are obtained by linear projection of the video content $f$ and the 3D poses into a common dimensional embedding space.

Finally, VPN is plugged into the 3D ConvNet for an end-to-end training with a regularized loss $L$ which is a convex combination of entropy loss, embedding loss and an attention regularization loss (refer to~\cite{VPN} for details).

 \begin{figure*}
\begin{center}
   \includegraphics[width=1\linewidth]{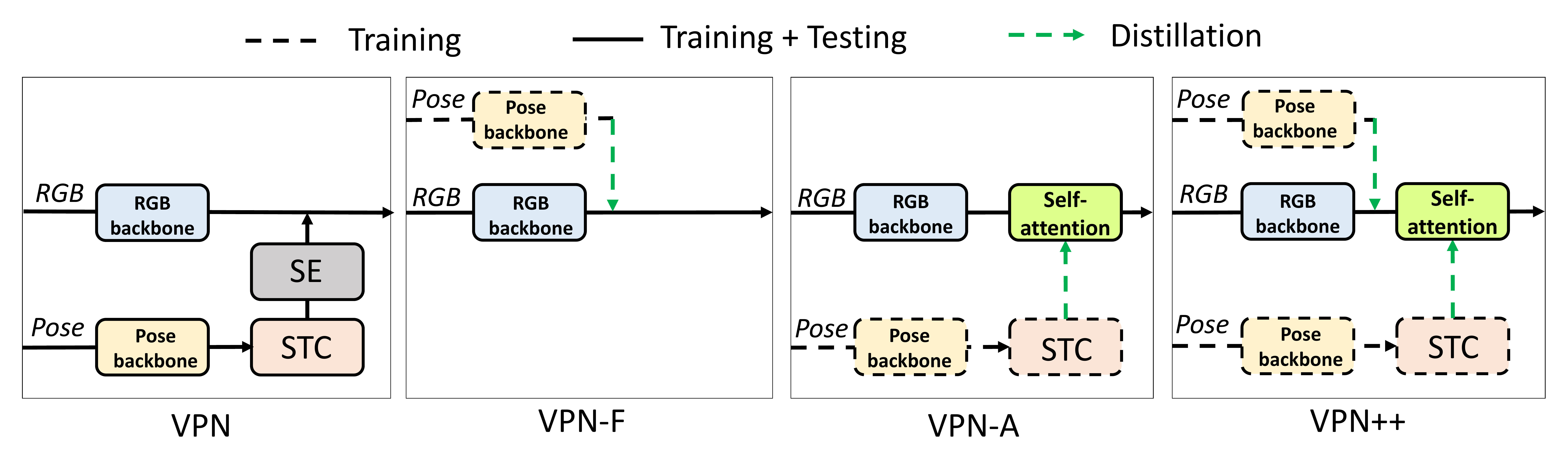} \vspace{-0.3in}
\end{center}
   \caption{A schematic diagram of our models - VPN, VPN-F, VPN-A, and VPN++ reflecting their variation for providing video-pose embeddings. The inputs to each model at training are the \textit{RGB} and \textit{Poses}. STC indicates the spatio-temporal coupler learning the attention weights. SE indicates the spatial embedding module.} 
\label{vpn_variants}
\end{figure*}

Aiming at learning similar video-pose embeddings by hallucinating discriminative pose-level features, we propose an extension of VPN, namely VPN++. VPN++ effectively makes use of the pose features at training time and eliminates its reliance over Poses at inference time. In fig.~\ref{vpn_variants}, we provide a schematic diagram of VPN and our proposed distillation models to illustrate the disparities among them. VPN++ with only feature-level distillation is denoted as VPN-F and VPN++ with only attention-level distillation is denoted as VPN-A.
Below, we elaborate the two levels of distillation in VPN++. 
\begin{figure*}
\begin{center}
   \includegraphics[width=1\linewidth]{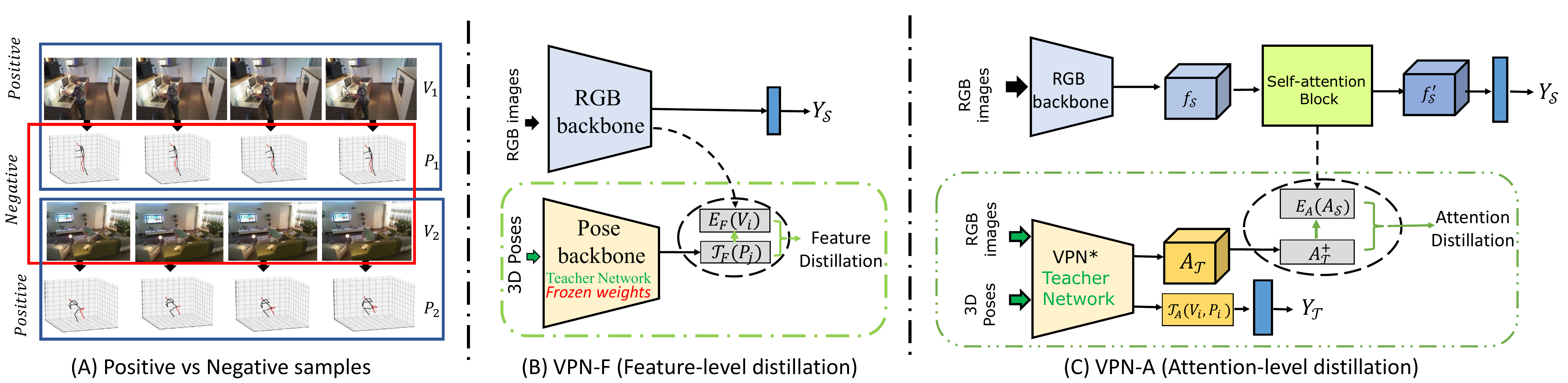}
\end{center}
   \caption{(A) The positive \& negative video-pose pairs (at the left) are input to the teacher-student network. (B) \textbf{VPN-F}: VPN++ distillation model with only feature-level distillation. Here, the Pose Teacher network is pre-trained for action classification. Supervised Contrastive Distillation (SCD) is applied between the RGB and Pose features. (C) \textbf{VPN-A}: VPN++ distillation model with only attention-level distillation. Here, the teacher VPN* is the video-pose network~\cite{VPN} without the spatial embedding (SE). Also, $A_\mathcal{T}$ and $\mathcal{T}_A(V_i,P_i)$ can be referred to as the attention weights ($A$) and modulated feature map ($f'$) of VPN (see fig.~\ref{vpn_fig}). Teacher network VPN* is trained collaboratively with the student RGB backbone.}
\label{model_1}
\end{figure*}
\subsection{VPN-F (Feature-level distillation)}
VPN++ involves knowledge distillation among modalities and thus, we have a teacher-student structure. This is referred to as the feature-level distillation in our model to infuse Pose stream into RGB stream. This is an attempt analogous to the spatial embedding in VPN.  
In order to perform this distillation, the Pose stream is considered as the Teacher Network $\mathcal{T}_F$, whereas the RGB stream as the Student Network $\mathcal{S}$.  
But unlike previous teacher-student networks~\cite{garcia_cross_modal_distillation, Luo_2018_ECCV}, here the teacher network occasionally provides irrelevant features, especially for actions where appearance information is important. For instance, using only Poses cannot discriminate actions like \textit{wearing a shoe} or \textit{taking off a shoe} but they can provide salient information about the localization of the action.
For disambiguating these actions with similar appearance, we have to go beyond just mimicking the Pose stream to capture discriminative information.
Consequently, we use the concept of contrastive learning to learn a representation for which the positive pairs are close to each other and negative pairs are pushed apart in some metric space. Most related to our work, Contrastive Representation Distillation~\cite{contrastive_distillation} (CRD) involves learning an unsupervised representation through knowledge distillation followed by a downstream training on the same set of training samples. 
In contrast, we focus specifically on video domain (with RGB and 3D poses) and formulate a supervised training strategy. This strategy includes jointly optimizing the student network with the class labels $\hat{Y}$ in addition to distilling the knowledge from Pose stream to RGB. This enables the actions with similar appearance to move apart in the feature space due to their dissimilar distillation through pose embeddings. We call the model with only this feature-level distillation as VPN-F.

\noindent At training time, we learn the VPN-F representation in two steps.
Let $V_i$ be a video (stack of RGB frames) and $P_i$ be the corresponding 3D Poses for the $i^{th}$ sample in the training set. In the first step, the teacher network $\mathcal{T}$ is trained with the 3D poses for classifying $V_i$ into $\mathcal{C}$ action classes and its weights are then frozen.

In the second step, our goal is to learn a latent space where semantically related RGB frames and Poses are close to each other and far away otherwise. We achieve this by imposing a supervised contrastive distillation (SCD) loss between the teacher and the student at the feature level as illustrated in fig.~\ref{model_1} (B).
Inspired from audio-video~\cite{owens2018audiovisual} and text-video analysis~\cite{miech20endtoend}, we assign a set of candidate positive pairs $(V_i, P_i)$, thus the RGB frames and 3D Poses are extracted from the same video labeled action $C_k \in \mathcal{C}$. On the other hand, the negative pairs are some randomly associated data $(V_i, P_j)$ where $P_j$ is randomly chosen from the subset $\mathcal{C} \setminus C_k$ as shown in fig.~\ref{model_1} (A).

\noindent For distillation, the SCD loss is imposed between the features at the output of the layer immediately before the final fully-connected layer of the teacher network and the features of
the visual embedding obtained from the RGB student network. This visual embedding $E_F(V_i)$ is a linear projection of $f_S$, where spatio-temporal feature map $f_S$ is computed by the RGB backbone $\mathcal{S}(V_i)$. 
We denote the features from the teacher network as $\mathcal{T}_F(P_j)$. 
We maximize the mutual information between Pose teacher and RGB student representations by jointly optimizing the student network at the same time as we learn a video-pose embedding $[\mathcal{T}_F(P_j),E_F(V_i)]$.
Thus, our distillation loss over a batch of data ($\mathcal{B}$) is formulated as the log likelihood of the data under this model:
\begin{equation}
\begin{aligned}
    \mathcal{L}_{SCD} = \frac{1}{|\mathcal{B}-\mathcal{N}|}\sum_{i}\log [\mathcal{T}_F(P_i),E_F(V_i)] + \\
     \sum_{j \neq i}\log (1 -[\mathcal{T}_F(P_j),E_F(V_i)]) \\
    \text{where} \hspace{10pt} [\mathcal{T}_F(P_j),E_F(V_i)] =  \frac{e^{\mathcal{T}_F(P_j)^{\intercal}E_F(V_i)}}{e^{\mathcal{T}_F(P_j)^{\intercal}E_F(V_i)} + \mathcal{M}}
\end{aligned}
\end{equation}
Here, $[\mathcal{T}_F(P_j),E_F(V_i)] \rightarrow (0, 1)$ corresponds to the video-pose embedding and constant $\mathcal{M}$ is determined by the ratio of the number of negatives $N$ to the cardinality of the dataset. Thus for the positive pairs, $\mathcal{L}_{SCD}$ enforces the video student representation $E_F(V_i)$ to project along the Pose teacher representation $\mathcal{T}_F(P_i)$. Conversely for the negative pairs, the student representation is projected perpendicular to the teacher representation in feature space. This feature modulation (at student network) due to the distillation loss is accompanied by cross-entropy loss $\mathcal{L}_C^{\mathcal{S}}$ to optimize the RGB student network for predicting the action labels $Y_S$. This joint optimization induces a selective infusion of the pose features in the RGB space with respect to the action class.
Note that the student network is always fed with the ground-truth $\hat{Y}$ corresponding to the RGB input. So, a video sample with corresponding ground-truth is repeated twice in a mini-batch while training VPN-F with SCD loss.

\subsection{VPN-A (Attention-level distillation)}
Next, we aim at learning RGB representation benefiting from attention mechanism. Attention mechanisms focusing on salient image region across time have become instrumental for discriminative visual representation.
For RGB based ADL recognition, VPN~\cite{VPN} has shown that pose driven attention mechanism is more accurate and effective compared to the ones using self-attention mechanisms through RGB itself. 
Therefore, we develop a second-level of distillation for transferring pose driven attention knowledge to RGB stream. For the sake of simplicity, we first explain the model with only attention-level distillation, dubbed as VPN-A.
For this distillation, we chose our Video-Pose Network~\cite{VPN} as a Teacher network $\mathcal{T}_A$. This Video-Pose Network (VPN*) is implemented following~\cite{VPN} with no spatial embedding since we find that the feature-level distillation could hallucinate the features learned through spatial embedding in VPN. 
 The student network $\mathcal{S}$ is a video backbone, similar to the one used in VPN-F.

\noindent The challenge is to transfer the knowledge of attention weights learned by the teacher to the RGB student network. Therefore, a self-attention block similar to~\cite{nonlocal} is invoked in the RGB based network which could learn the attention weights from the teacher. However, a feature-level distillation in this case does not activate the relevant neurons at the student network. We empirically support this claim in the experimental analysis. Moreover, learning attention weights is an evolutionary mechanism where a model learns the salient regions in the spatio-temporal space with every batch of iteration over the training data. So, for this level of distillation, we opt for online distillation, where the teacher VPN* and the student RGB backbone along with the self-attention block collaboratively optimize their respective entropy loss as illustrated in fig.~\ref{model_1} (C).
Such a distillation encourages the RGB student to produce similar attention weights as the VPN* teacher, intuitively paying attention to similar parts of the video as the teacher.

\begin{figure}
\begin{center}
   \includegraphics[width=1\linewidth]{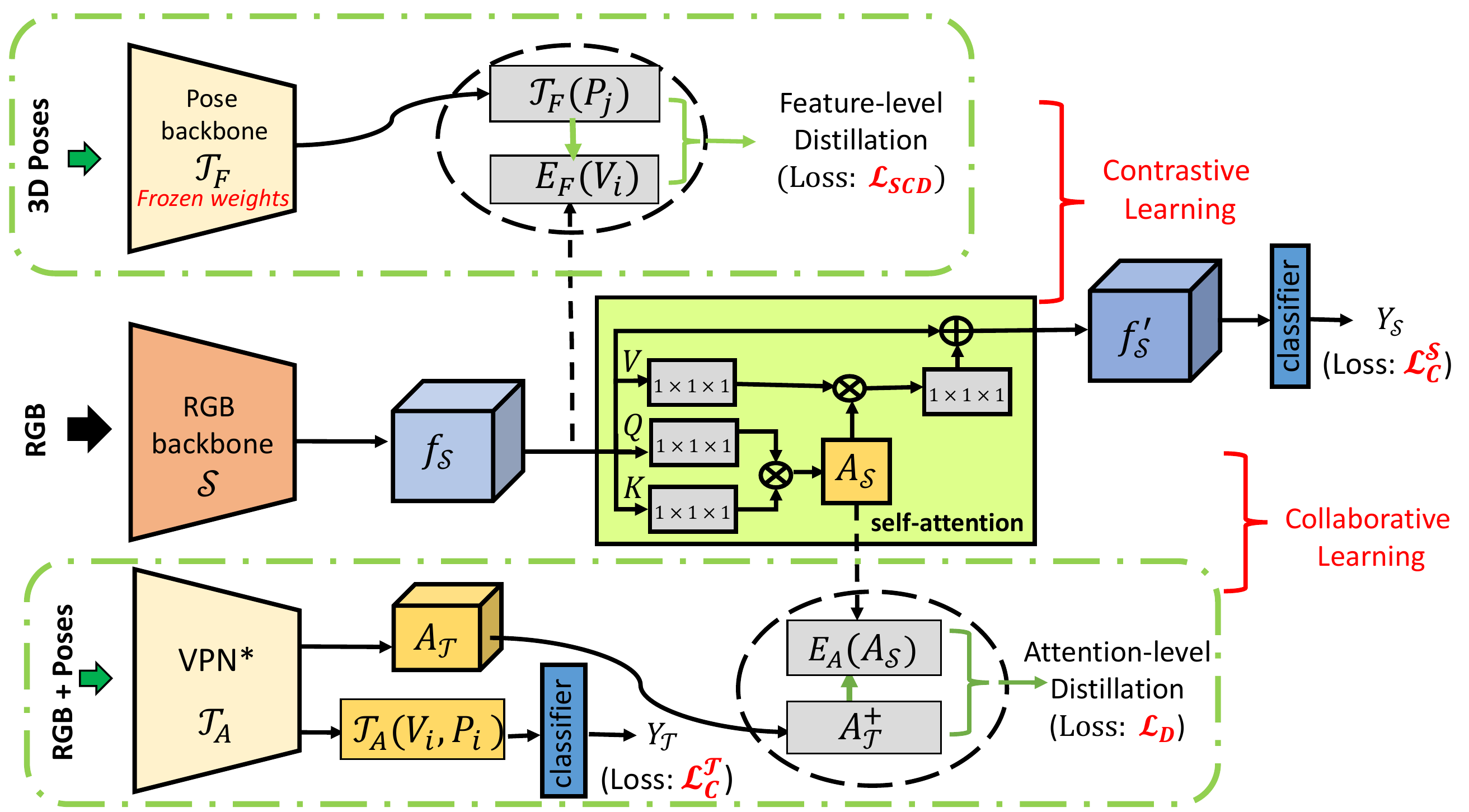}
\end{center}
   \caption{\textbf{VPN++}: The proposed distillation model when both VPN-F and VPN-A are integrated into a single model. The student  network consists of a RGB backbone and a self-attention bock. At training, the model is trained in a contrastive manner for the feature-level distillation, and collaborative manner for the attention-level distillation. Note that Video-Pose attention model VPN* does not have the spatial embedding module.}
\label{model_integrated}
\end{figure}

\noindent VPN-A is trained in a single step. On one hand, VPN* teacher network is trained with action labels to learn pose driven attention weights $A_{\mathcal{T}}$ to modulate its RGB feature map. Note that these attention weights corresponds to $A$ from STC of VPN discussed in section 3.1.
On the other hand, the student network intakes only the RGB frames. The self-attention block projects the RGB feature $f_{\mathcal{S}}$ to a query ($Q$) and memory (key and value, $K$ \& $V$ ) embedding using linear projections ($1 \times 1 \times 1$ Conv), where typically the query and keys are of lower dimension (see fig.~\ref{model_integrated} for a zoom into the self-attention block). The output for the query, i.e. the modulated feature map $f'_{\mathcal{S}}$, is computed as an attention weighted sum of values $V$, with the attention weights $A_\mathcal{S}$ obtained from the product of the query $Q$ with keys $K$. The attention weights $A_\mathcal{S}$ have to be learned from the evolution of 3D Poses. So, we invoke a distillation loss between the self-attention and VPN attention weights ($A_\mathcal{S}$ \& $A_\mathcal{T}$).
The distillation loss is a classical Mean Squared Error (MSE) loss between the attention weight embeddings ($E_A(A_{\mathcal{S}})$ \& $A^+_{\mathcal{T}}$) from the teacher-student network. The projection of the attention weights ($A_{\mathcal{S}}$ \& $A_{\mathcal{T}}$) is necessary since they differ in terms of their dimensionality. This projection is performed by linearly transforming them into the same dimension and further normalizing them through $\mathcal{L}$-2 norm. The distillation loss $\mathcal{L_D}$ is formulated as 
\begin{equation}
    \mathcal{L}_D = ||A_{\mathcal{T}}^{+} - E_A(A_{\mathcal{S}})||^2
\end{equation}
The VPN* backbone, i.e. $\mathcal{T}_A(V_i, P_i)$ classifies its modulated feature map using the entropy loss $L_C^{\mathcal{T}}$ between the true class labels $\hat{Y}$ and the predicted class labels $Y_{\mathcal{T}}$.
Besides, the modulated feature map $f'_{\mathcal{S}}$ at the student network is classified simultaneously using the entropy loss $L_C^{\mathcal{S}}$ between the same true class labels and the predicted class labels $Y_{\mathcal{S}}$.  

\subsection{VPN++: Integrating VPN-F \& VPN-A}
Finally, we aim at learning a unified RGB representation that can emulate both Pose based features and pose driven attention weights.
This objective also encourages the model to jointly optimize the two levels of distillation loss along with the cross entropy loss to learn the class labels.
Thus, we integrate the two levels of distillation into a single model - we call VPN++. The training methodology of VPN++ involves contrastive learning for the feature-level distillation and collaborative learning for the attention-level distillation. In fig.~\ref{model_integrated}, we show the VPN++ model with two levels of distillation. Here the RGB student network includes the RGB backbone and the self-attention block whereas there are two teacher networks - a pre-trained Pose backbone for infusing the pose features to the RGB stream, and VPN* for transferring the pose driven attention knowledge to the self-attention block of the student network. In order to incorporate the contrastive and collaborative learning strategies both in the same model, a batch of samples with (positive, negative) pairs for the feature-level distillation and (positive, positive) pairs for the attention-level distillation is fed to the model. Note that the Pose teacher network for feature-level distillation is frozen.

\noindent Thus, the RGB student network is jointly optimized with the following linear combination of the distillation losses and the entropy losses:
\begin{equation}
    \mathcal{L} = \mathcal{L}_C^{\mathcal{S}}(Y_S, \hat{Y}) + \mathcal{L}_C^{\mathcal{T}}(Y_{\mathcal{T}}, \hat{Y}) - \alpha \mathcal{L}_{SCD} + \beta \mathcal{L}_{D} 
\end{equation}
where $\alpha$ and $\beta$ are the weighting factors of the distillation losses. Thus, VPN++ not only learns to distill the pose knowledge into RGB but also learn discriminative representation through pose driven attention distillation. 
While testing VPN++ (the RGB student network), we only use RGB frames as input to compute the action class scores, avoiding the requirement of 3D Poses.

\section{Experiments}
We evaluate the effectiveness of VPN++ and its corresponding components for action classification on four datasets popular for ADL: a real-world dataset - Toyota-Smarthome~\cite{STA_iccv}, a large scale human activity dataset - NTU RGB+D-60~\cite{NTU_RGB+D}, the super-set of NTU-60 dataset - NTU RGB+D-120~\cite{ntu120}, and a relatively small scale human-object interaction dataset - Northwestern-UCLA~\cite{nucla}.  

\noindent\textbf{Toyota-Smarthome} (Smarthome or SH) is a recent ADL dataset recorded in an apartment where 18 older subjects carry out tasks of daily living during a day. The dataset contains 16.1k video clips, 7 different camera views and 31 complex activities performed in a natural way without strong prior instructions. This dataset provides RGB data and 3D skeletons which are extracted from LCRNet~\cite{lcrnet_new}. For evaluation on this dataset, we follow cross-subject ($CS$) and cross-view ($CV_2$) protocols proposed in~\cite{STA_iccv}. We ignore protocol $CV_1$ due to limited training samples. 
 
\noindent\textbf{NTU RGB+D} (NTU-60 \& NTU-120): NTU-60 is acquired with a Kinect v2 camera and consists of 56880 video samples with 60 activity classes. The activities were performed by 40 subjects and recorded from 80 viewpoints. For each frame, the dataset provides RGB, depth and a 25-joint skeleton of each subject in the frame. For evaluation, we follow the two protocols proposed in~\cite{NTU_RGB+D}: cross-subject (CS) and cross-view (CV). NTU-120 is a super-set of NTU-60 adding a lot of new similar actions. NTU-120 dataset contains 114k video clips of 106 distinct subjects performing 120 actions in a laboratory environment with 155 camera views. For evaluation, we follow a cross-subject ($CS_1$) protocol and a cross-setting ($CS_2$) protocol proposed in~\cite{ntu120}.

\noindent\textbf{Northwestern-UCLA Multiview activity 3D Dataset} (N-UCLA) is acquired simultaneously by three Kinect v1 cameras. The dataset consists of 1194 video samples with 10 activity classes. The activities were performed by 10 subjects, and recorded from three viewpoints. We performed experiments on N-UCLA using the cross-view (CV) protocol proposed in~\cite{nucla}: we trained our model on samples from two camera views and tested on the samples from the remaining view. For instance, the notation $V_{1,2}^3$ indicates that we trained on samples from view 1 and 2, and tested on samples from view 3. 

\subsection{Implementation details}
For the input at training time, the 3D Poses are provided for NTU and N-UCLA dataset.
For Smarthome dataset, two sets of 3D Poses, namely old and new Poses, are provided which are eventually extracted from RGB. Note that the new 3D Poses are of higher quality compared to the older ones.

\noindent \textbf{For VPN++}, the Teacher network for  feature-level distillation is AGCN-J~\cite{2sagcn2019cvpr} Pose backbone. Thus, we follow the pre-processing step on the 3D Poses as in~\cite{2sagcn2019cvpr}. For attention-level distillation, the Teacher Network (VPN*) is adapted with a 2 layer AGCN as Pose backbone and no spatial embedding.
The Student network is I3D~\cite{i3d} RGB backbone pre-trained on ImageNet~\cite{imagenet_cvpr09} and Kinetics-400~\cite{kinetics}. It takes 64 RGB frames as input. 
The self-attention block is implemented with an additional Non-Local block~\cite{nonlocal} placed on top of the I3D (\texttt{Mixed\_5c} layer).

\noindent \textbf{Training}. For training the teacher networks of VPN++ with categorical cross entropy loss, we follow the steps as in~\cite{2sagcn2019cvpr} and~\cite{VPN}. For training the student network, a dropout~\cite{Dropout} of 0.3 and a \textit{softmax} layer are added at the end of the self-attention block for class prediction. VPN++ is trained with a 4-GPU machine where each GPU has 4 video clips in a mini-batch. It is trained with SGD optimizer having initial learning rate of 0.01, momentum of 0.9, and a weight decay rate of 0.1 after every 10 epochs. While training VPN++, we chose $\alpha = \beta = 50$. For feature-level distillation, each batch consists of 8 positives and 8 negatives. 

\noindent \textbf{Inference.} At test time, we perform fully convolutional inference in space as in~\cite{nonlocal}. The final classification is obtained by max-pooling the softmax scores.

\subsection{Hyper-parameter sensitivity}
\begin{figure}
\begin{center}
   \scalebox{1.0}{\includegraphics[width=1\linewidth]{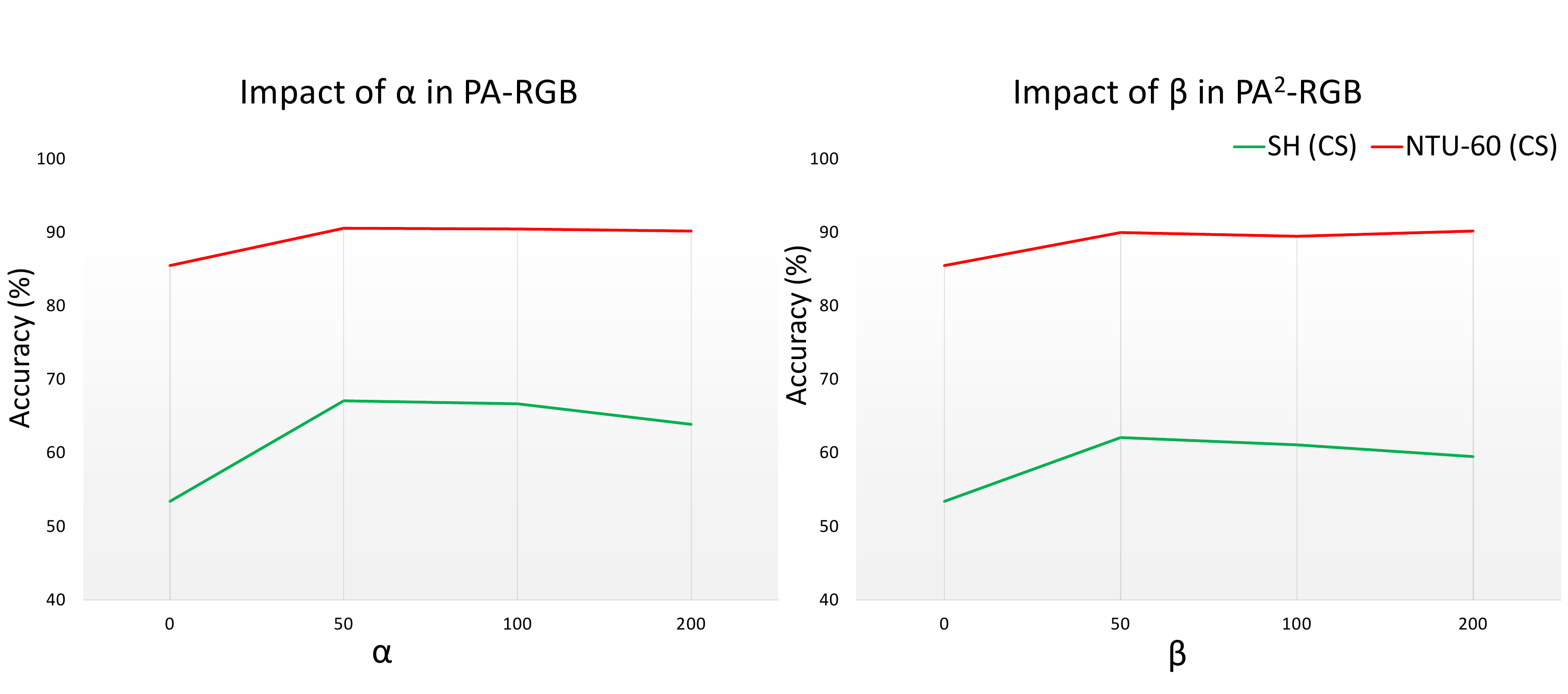}}
\end{center}
   \caption{Accuracy of VPN-F (on left) and VPN-A (on right) for different values of $\alpha$ \& $\beta$ respectively on Smarthome (CS) and NTU-60 (CS) datasets.}
\label{alpha}
\end{figure}
VPN-F and VPN-A distillation models are trained by a linear combination of two losses: cross-entropy loss between the logits and the ground-truth targets, and the distillation loss between the video-pose features. In fig.~\ref{alpha}, we report the accuracy of VPN-F and VPN-A on Smarthome and NTU-60 datasets using different values of $\alpha$ and $\beta$ respectively. We observe that a non-zero value of $\alpha$ or $\beta$ increases the action classification accuracy compared to the baseline RGB stream. This shows the importance of the distillation from Pose stream to RGB in both models. We also observe that by increasing the weighting factor of the distillation loss, we reach a peak accuracy for $\alpha = \beta = 50$ as shown in fig.~\ref{alpha}. This shows that our distillation models effectively leverage both RGB and Pose streams to classify the action when combined in a strategic manner. Further increase in the values of $\alpha$ or $\beta$ influences the distillation loss to dominate the student RGB network while training. This causes the resultant student network to mimic the Pose stream rather than exploiting both streams.

\noindent For SCD loss, the choice of number of negatives for each positive input (video-pose pair) is flexible. Conventionally, more negatives for each Positive in contrastive learning yields higher accuracy. This observation is not noted in our case due to a supervised strategy of using the contrastive loss. Thus, we take one negative for each positive to train VPN-F. We utilize the above observation for hyper-parameters while training VPN++. 
\subsection{Ablation studies}
In this section, we analyze the impact of proposed distillation methods w.r.t. previous methods. We also quantify the robustness of VPN-F and VPN-A.
\begin{table}[t]
  \centering
\caption{Ablation for choice of distillation loss in VPN-F.}{%
\scalebox{0.9}{
\begin{tabular}{l|ccc}
\hline
Loss  & SH   & NTU-60  & NTU-60 \\
 & (CS)  & (CS) & (CV)    \\
\hline
MSE~\cite{mars} &  61.8  & 89.1 & 92.4 \\
CRD~\cite{contrastive_distillation} & 64.7 & 90.1 & 93.1 \\
SCD &  \textbf{67.1} & \textbf{90.8} & \textbf{93.8} \\
\hline
\end{tabular}}
}
\label{ablation_scd}
\end{table}
\begin{table}[t]
  \centering
\caption{Impact of pose driven attention (VPN) compared to RGB based Non Local (NL) attention mechanism. }{%
\scalebox{0.8}{
\begin{tabular}{lc|ccc}
\hline
Model & Poses & SH    & NTU-60  & NTU-60 \\
      &       & (CS)  & (CS)    & (CV)    \\
\hline
I3D w/o attention (backbone) & $\times$ &  53.4 & 85.5 & 87.3  \\
I3D w NL attention (self-attn) & $\times$ & 53.6 & 88.4 & 87.1 \\
I3D w pose attention (VPN) & \checkmark  & \textbf{65.2} & \textbf{93.5} & \textbf{96.2} \\
\hline
\end{tabular}}
}
\label{ablation_attention}
\end{table}
\begin{table}[t]
  \centering
\caption{Comparison of VPN-A with other strategies to distill pose driven attention.}{%
 \scalebox{0.8}{
\begin{tabular}{lc|ccc}
\hline
 Loss \& distillation & \small{Collaborative} & SH    & NTU-60  & NTU-60 \\
    strategies     & learning  & (CS)  & (CS)    & (CV)    \\
\hline
SCD (feature level) & $\times$ & 55.7  & 89.1 & 91.5  \\
SCD (feature level) & \checkmark & 51.1  & 87.1 & 89.5  \\
SCD (attention weights)& $\times$ & 54.2 & 88.9 & 91.4 \\
SCD (attention weights)& \checkmark & 53.1 & 87.4 & 90.6 \\
MSE (attention weights)& $\times$ &  61.1 & 89.1 & 92.4   \\
MSE (attention weights)& \checkmark &   \textbf{62.1} & \textbf{90.0} & \textbf{93.1}   \\
\hline
\end{tabular}
}
\label{ablation_att_dist}}
\end{table}

\begin{table}[t]
  \centering
\caption{Comparison of different strategies to combine VPN-F \& VPN-A. $L_e$ represents the spatial embedding in VPN* (teacher of VPN-A). }{%
\scalebox{0.8}{
\begin{tabular}{c|c|c|ccc}
\hline
Combining strategy & $L_e$ & Test & SH & NTU-60 & NTU-60 \\
        & & time (s) & CS   & CS & CV \\
\hline
VPN-F ($1^{st}$) + VPN-A ($2^{nd}$) & $\times$ & 0.4 & 62.3 & 90.5 & 93.2 \\
VPN-A ($1^{st}$) + VPN-F ($2^{nd}$)& $\times$&  0.4 & 63.9 & 90.6  & 93.4 \\
\hline
VPN-F + VPN-A (Late Fusion) & $\times$& 0.7 & 68.7 & 91.7 & 94.8 \\
VPN++ (multi-teacher) & \checkmark & 0.4 & 68.9 & \textbf{91.9} & 94.8 \\
VPN++ (multi-teacher)& $\times$ & 0.4 & \textbf{69.0} & \textbf{91.9} & \textbf{94.9} \\
\hline
\end{tabular}
}
\label{pose_quality_1}
}
\end{table}

\begin{table}[t]\tabcolsep=1.1pt
  \centering
\caption{Top-1 accuracy of RGB, 3D Poses, VPN-F, VPN-A, and VPN++ on 4 datasets. }{%
\scalebox{0.8}{
\begin{tabular}{ll|cccccc}
\hline
& Stream & SH & NTU-60 & NTU-60 & NTU-120  & NTU-120  & N-UCLA  \\
 &      &  (CS)  &  (CS)      &  (CV)       & ($CS_1$) & ($CS_2$) & ($V^3_{1,2}$)    \\
\hline
$l_1$: & RGB  & 53.4 & 85.5 & 87.3 & 77.0 & 80.1 & 86.0 \\
$l_2$: & 3D Poses & 51.5 & 85.8 & 93.8 & 79.6 & 81.1 & 78.2 \\
& $l_1 + l_2$ (\small{Late Fusion})   & 63.0  & 87.7 & 94.8 & 81.1 & 83.3 & 87.1\\
& $l_1 + l_2$ (\small{attention})   & 65.2  & 93.5 & 96.2 & 86.3 & 87.8 & 93.5 \\
\hline
{\multirow{5}{*}{\rotatebox{90}{Ours}}} & VPN-F & 67.1 & 90.8  & 93.8 & 85.1 & 87.6 & 89.1\\
& VPN-A  & 62.1  & 90.0 & 93.1 & 85.2 & 88.0 & 88.2\\
& VPN++  & 69.0 & 91.9  & 94.9 & 86.7 & 89.3 & 91.9\\
& VPN++ + 3D Poses & \textbf{71.0} & \textbf{94.9}  & \textbf{98.1} & \textbf{90.7} & \textbf{92.5} & \textbf{93.5} \\
\hline
\end{tabular}
}
\label{pose_rgb}
}
\end{table}

\begin{table}[t]
  \centering
\caption{Performance  of  several methods with different levels of pose quality.}{%
\begin{tabular}{cc|ccc}
\hline
Dataset  & Pose    & AGCN-J & VPN & VPN++ \\
         & Quality   & \cite{2sagcn2019cvpr}& \cite{VPN} &  \\
\cline{1-5}
SH (CS) & Medium & 54.0 & 65.2 & \textbf{69.0} \\
SH (CS) &  Low  & 49.1 & 62.1 & \textbf{66.8} \\
\cline{1-5}
NTU-60 (CS) & High & 85.8 & \textbf{93.5} & 91.9 \\
NTU-60 (CS) & Low & 44.4 & 90.1 & \textbf{91.3} \\
\hline
\end{tabular}
}
\label{pose_quality_2}
\end{table}
\noindent \textbf{Which loss is better for feature-level distillation?} In this ablation study (Table~\ref{ablation_scd}), we compare different distillation losses for transferring knowledge from pose features to RGB features. The training strategy for all these losses are different but are applied between the video-pose features $\mathcal{T}_F(P_j)$ and $E_F(V_i)$. The mechanism of learning visual representation with the concept of contrastive learning between the positive and negative samples (CRD~\cite{contrastive_distillation} and SCD) outperform the classical way to distillate knowledge using MSE loss~\cite{mars}. 
We also note that our SCD outperforms CRD significantly on Smarthome, whereas the margin of improvement on NTU is comparatively low. This indicates that CRD is effective for scenarios where high quality Poses are available and SCD is consistently effective even for low quality Poses. 

\noindent \textbf{Why do we need to emulate pose driven attention?} We know that attention mechanisms are crucial for understanding ADL~\cite{STA_iccv}. But attention weights obtained using RGB based self-attention mechanism like Non-Local blocks~\cite{nonlocal} rely too much on variation of intensities in spatio-temporal feature maps, hence lacks semantics. In contrast, 3D poses capture the semantics in the videos and significantly improve the action recognition performance as shown in Table~\ref{ablation_attention}. Our VPN with pose driven attention significantly improves the action classification accuracy by relatively 20.8\% on Smarthome dataset. It is worth noting that the improvement is significant for Smarthome compared to NTU-60 as it contains many fine-grained actions with videos captured by fixed cameras in an unconstrained Field of View. Thus, enforcing  the  embedding loss enhances the spatial  precision during inference.
Therefore, we chose to mimic pose driven attention for a second-level of distillation in VPN++.

\noindent \textbf{Which loss is better for attention-level distillation?} 
For VPN-A, we propose to distill knowledge at attention-level than at feature-level due to its more effectiveness as supported by the experiments in Table~\ref{ablation_att_dist}. Note that the feature-level distillation in the former experiment is performed between the output of the modulated feature map of VPN*, i.e. $\mathcal{T}_A(V_i, P_i)$ and the modulated feature map of the student network $f'_\mathcal{S}$.
We investigate the effectiveness of collaborative training the teacher-student network for transferring attention-level features. In these experiments in Table~\ref{ablation_att_dist}, the VPN* teacher network is pre-trained and frozen when collaborative training is not performed. 
We also compare the performance of supervised contrastive distillation loss (most effective loss in Table~\ref{ablation_scd}) with MSE loss at attention-level. However, MSE loss with collaborative training strategy to distill attention weights from VPN* Teacher network to RGB based Non-Local student network outperforms the baselines by up to 7.9\% on Smarthome dataset. This shows that reducing MSE between the attention weights of video and pose embeddings is a better strategy to distill attention weights than contrastive learning. This is coherent with the fact that distillation of attention weights do not correspond to positives and negatives w.r.t. video samples whereas distillation at feature level represents entities that could have positives and negatives. 

\noindent \textbf{How to combine VPN-F \& VPN-A?} In Table~\ref{pose_quality_1}, we observe a performance drop when both the training strategies of VPN-F \& VPN-A are combined in a sequential manner, one after the other. The cause for this performance drop is the difficulty for the second distillation to significantly modify the RGB feature map (at student's network) once the first distillation has modified it.
In contrast to these fusion strategies, the score level fusion of both student networks significantly outperforms the above two end-to-end strategies. Similar improvement is noted for our multi-teacher network trained with contrastive and collaborative strategy. Due to the lower final model complexity and lower inference time, the multi-teacher network is superior than the one with late fusion.
This performance improvement highlights the complementary optimizations which are well preserved while training jointly in a single model in VPN++. Finally, we also observe that spatial embedding in the teacher network of attention-level distillation do not contribute to the classification accuracy and hence can be ignored. This shows that the feature-level distillation could hallucinate the pose features performed by the spatial embedding in VPN.

\noindent \textbf{Comparison of distilled models with RGB \& Pose streams.} In Table~\ref{pose_rgb}, we compare our distillation models - VPN-F \& VPN-A with uni-modal models and their combinations. RGB and 3D Poses are modeled using I3D~\cite{i3d} and AGCN-J~\cite{2sagcn2019cvpr} networks. Following the state-of-the-art trends, RGB and Poses are combined using score level fusion (Late Fusion) and attention mechanism (VPN). Both VPN-F \& VPN-A significantly outperform the individual modalities. VPN-F with contrastive distillation outperforms the late fusion strategy of combining RGB and Poses on all the datasets except NTU-60 (CV protocol). This exception is coming from the high action classification performance ($93.8\%$) with view-invariant 3D poses for cross-view protocol of NTU-60, where high quality 3D Poses are available. On the other hand, VPN-A is an attention based model and requires subsequently large amount of data for learning salient attention weights. This is corroborated by its lower classification accuracy for NTU-60 in contrast to NTU-120 (up to $0.4\%$ higher than even VPN-F) where the number of training samples is two times that of NTU-60. The combination of VPN-F \& VPN-A in VPN++ further boosts the classification accuracy by up to $2.8\%$ relatively on Smarthome. Further improvement in action classification when combined with 3D Poses indicates that our distilled models still lacks in terms of mimicking the Pose stream. However, their superior performances compared to all prior techniques of combining RGB and 3D Poses show the discriminative representation learned by our VPN++.

\noindent \textbf{What happens when the Pose quality degrades?} As shown in fig.~\ref{time_analysis_intro}, VPN++ does not require Poses at test time which substantially reduces the model inference time. Thus, the bad quality of Poses at inference time does not hamper the performance of these models. But what if the Poses are shoddy for the entire data distribution (even at training)?  
In Table~\ref{pose_quality_2}, we dig deeper into this problem by investigating the influence of Pose quality on the performance for different models. First, we describe the experimental setup to obtain the different levels of Pose quality.

\noindent NTU-60 dataset was recorded in a laboratory, so we can have \textbf{hight-quality 3D Poses} captured by the Microsoft Kinect v2 sensor. 
For \textbf{low-quality 3D Poses}, we down-scaled the original videos by reducing their resolution to $320\times180$ and randomly invoke partial occlusions to fabricate the dataset similar to real-world settings. Then, we extract the 3D Poses using LCRNet++~\cite{lcrnet_new}. In Smarthome, Poses are obtained from RGB rather than using depth-map. For \textbf{medium-quality 3D Poses}, we apply Selective Spatio-Temporal Aggregation based Pose Refinement System (SSTA-PRS)~\cite{yang2020selective} which aims at improving the performance of pose estimation by integrating the advantages of several state-of-the-art pose estimation systems (eg. LCRNet++~\cite{lcrnet_new}, OpenPose~\cite{OpenPose} and AlphaPose~\cite{fang2017rmpe}) to extract 2D Poses. Then, we apply VideoPose3D~\cite{videopose3d} to obtain 3D Poses over 2D Poses. For the \textbf{low-quality 3D Poses}, we only use LCRNet++~\cite{lcrnet_new}.

\noindent The two baselines compared with VPN++ in Table~\ref{pose_quality_2} include skeleton based model: AGCN-J~\cite{2sagcn2019cvpr} and RGB+Pose based attention model: VPN. We observe that VPN++ is less sensitive to the quality of Poses with a deterioration of classification accuracy by 3.1\% and 0.6\% on Smarthome and NTU-60 respectively compared to the baselines (4.7\% \& 3.6\% for VPN and 9\% \& 48.2\% for AGCN-J). This experiment shows that the quality of Poses highly impacts skeleton based action recognition, whereas our distillation model VPN++ outperforms even VPN without the requirement of Poses at inference.
The tolerance of VPN++ to noisy Poses is due to the selective distillation of Pose features within the video-pose embedding of the distillation mechanisms. For instance, the degraded Poses contingent upon occlusions, low subject resolution, or other real world scenarios provide ambiguous features pertaining to actions. Thanks to the distillation mechanisms, we can filter the appropriate pose based features while infusing knowledge into the RGB stream.

%
\vspace{-0.08in}
\section{Qualitative Visualization}
\begin{figure}
\begin{center}
   \scalebox{1.05}{\includegraphics[width=1\linewidth]{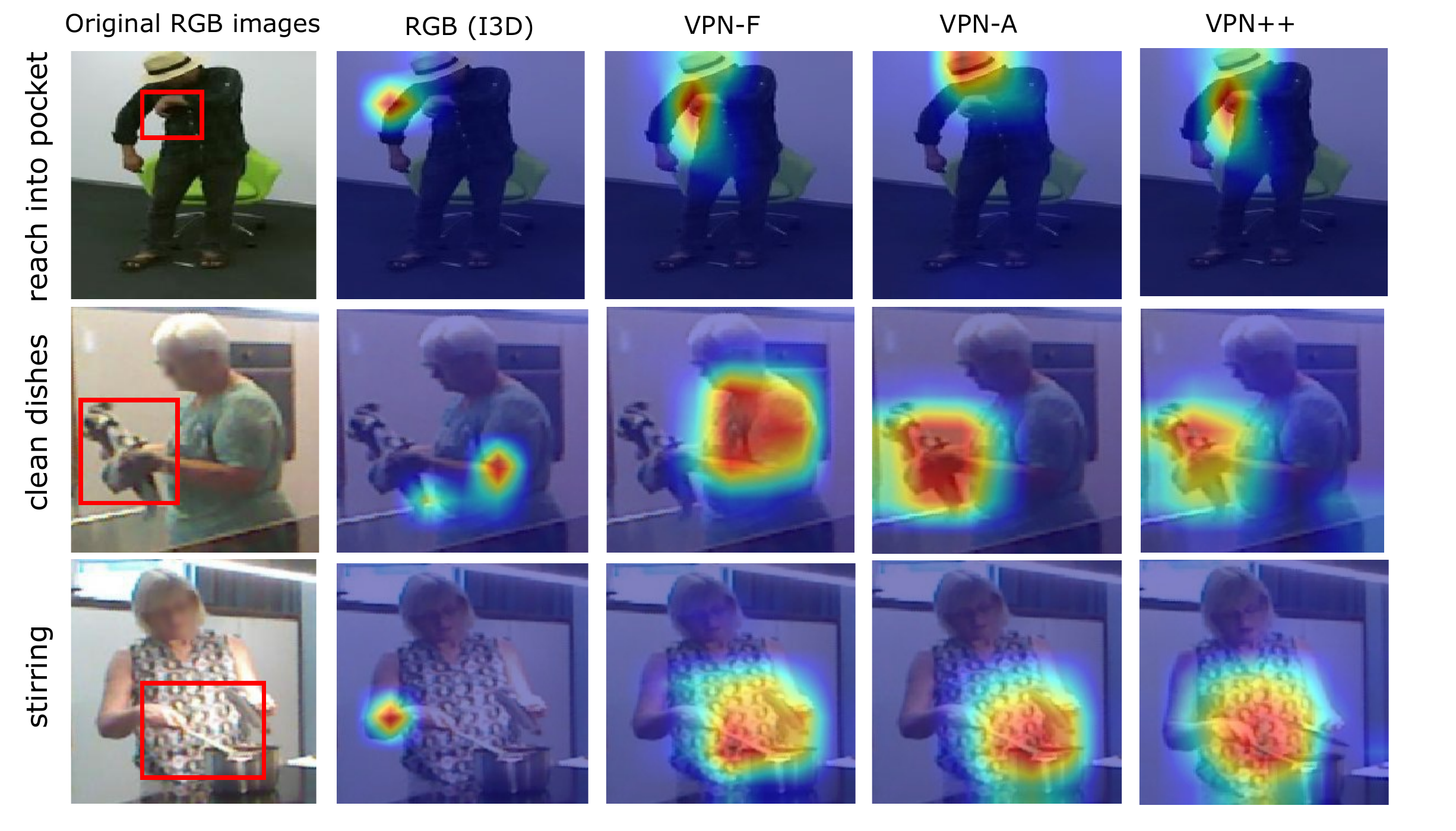}}
\end{center}
   \caption{Qualitative visualization of class activation maps of RGB, VPN-F, VPN-A, and VPN++ using Grad-CAM~\cite{grad_cam}. The \textcolor{red}{red} bounding box refers to the precised Region of Interest relevant to classifying the action.}
\label{viz}
\end{figure}

In fig.~\ref{viz}, we present a visualization of class activation maps of RGB, VPN-F, VPN-A, and VPN++ using Grad-CAM~\cite{grad_cam}. These maps enable us to visualize discriminative regions specific to each action class. VPN-F, for actions like \textit{reach into pocket}, VPN-A for actions like \textit{clean dishes}, and both VPN-F \& VPN-A for actions with subtle motion like \textit{stirring}, focus sharply  around the hands grasping objects providing contextual information worth modeling the actions.
The activation map of RGB stream either focuses only on irrelevant motion patterns (see fig.~\ref{viz}). Moreover, the class activation maps obtained for VPN++ select the ones (VPN-F or VPN-A) that is effective for an action class. Thus, our combining strategy of the two levels of distillation in VPN++ can take advantage of the complementary features learned via both distillation mechanisms. 
This qualitative visualization shows that our distillation mechanisms learn discriminative representation that exploits the contextual information in the scene which is crucial for ADL. 
\section{Quantitative Analysis}
\begin{figure}
\begin{center}
   \scalebox{1.0}{\includegraphics[width=1\linewidth]{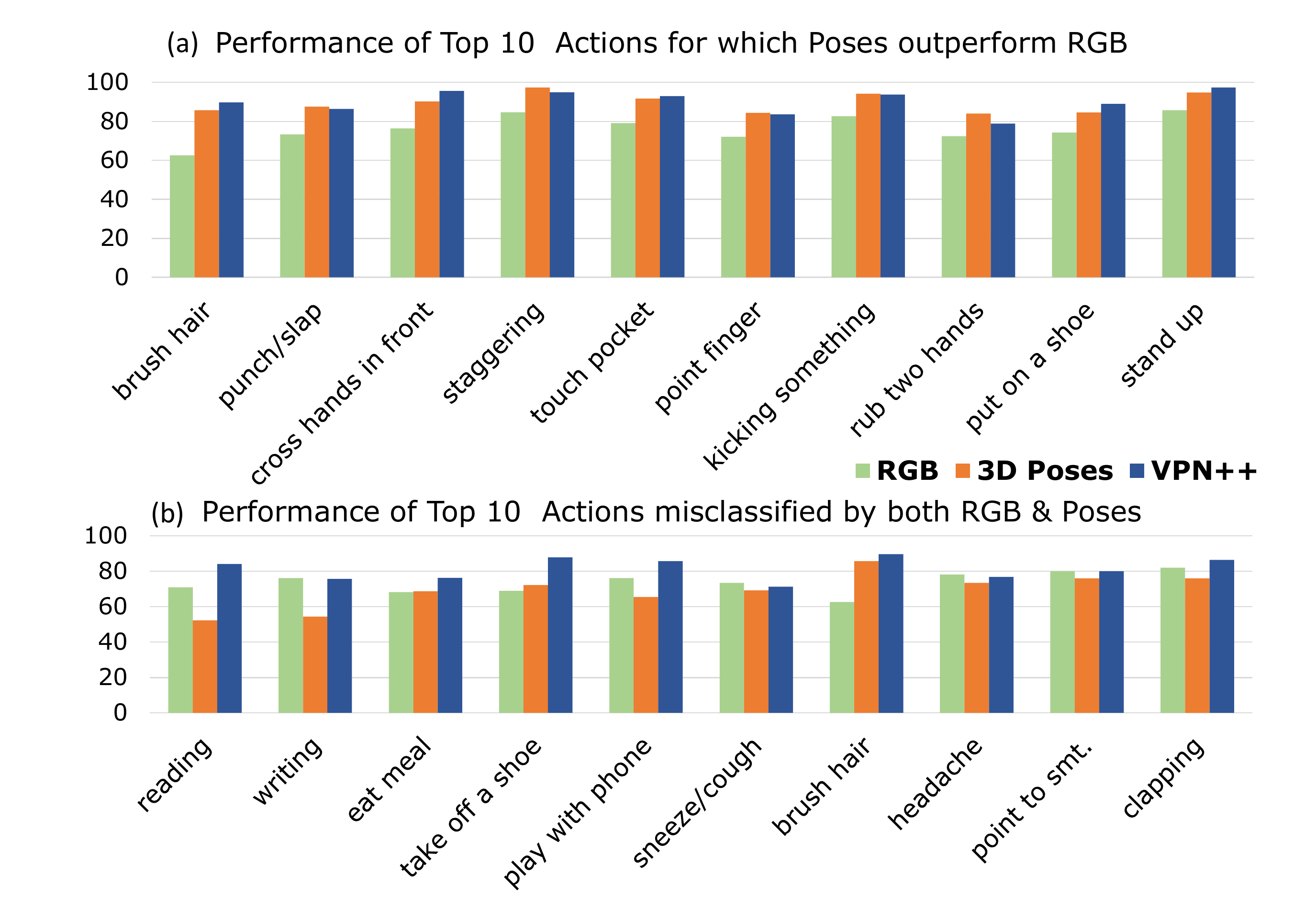}}
\end{center}
   \caption{Action classification accuracy (in \%) of top-10 actions (a) for which Pose stream outperforms RGB stream, and (b) which are mis-classified by both RGB and Poses.}
\label{graph}
\end{figure}
In this section, we present an analysis of Top-10 class-wise performance of an RGB based approach, a Pose based approach, \& our distillation model VPN++ (see fig.~\ref{graph}). First, (a) we present the performance of Top 10 actions for which Poses outperform RGB. VPN++ outperforms the Pose stream for all these actions which shows that our model not only learns Pose based features but also learns an augmented representation from the Pose teacher network. Second, (b) we present the performance of Top 10 actions mis-classified by both RGB and Pose streams. Interestingly, VPN++ improves the performance of RGB stream for actions which are mostly mis-classified owing to two challenges - (i) similarity in appearance like \textit{taking off a shoe} (+5\%) or \textit{wearing a shoe}, \textit{clapping} (+4\%) or \textit{rubbing two hands}, and (ii) subtle motion while performing the actions like \textit{reading} (+13\%), \textit{writing} (+1\%), and \textit{headache} (+2\%). Thus, VPN++ confirms empirically its potential to mitigate the drawbacks of SoA approaches by effectively providing an appropriate combination of the modalities (RGB and Poses) through distillation.

\begin{table}[t]
  \centering
\caption{Results on Smarthome dataset with cross-subject (CS) and cross-view ($CV_2$) settings (accuracies in \%); Att indicates attention mechanism, $\circ$ indicates that the modality has been used only in training.}{
\scalebox{0.925}{
\begin{tabular}{llccccc}
\hline
&Methods & Pose & RGB & Att & CS &  $CV_2$  \\
\hline\hline
&DT~\cite{DT} & $\times$ & \checkmark & $\times$ & 41.9 &  23.7 \\
&I3D~\cite{i3d} & $\times$ & \checkmark & $\times$ &53.4  & 45.1 \\
&I3D+NL~\cite{nonlocal} & $\times$ & \checkmark & \checkmark &  53.6 &  43.9 \\
&AssembleNet$++$~\cite{assemblenetplusplus} & $\times$ & \checkmark & \checkmark & 63.6 & -  \\
\hline
{\multirow{5}{*}{\rotatebox{90}{Old Poses}}}&LSTM~\cite{lstm3d} & \checkmark & $\times$ & $\times$ & 42.5 &  17.2 \\
&P-I3D~\cite{spatial-i3d} &\checkmark &\checkmark &\checkmark& 54.2 &50.3\\
&Separable STA~\cite{STA_iccv} & \checkmark & \checkmark & \checkmark & 54.2  &  50.3  \\
&VPN~\cite{VPN}  & \checkmark & \checkmark &\checkmark  & 60.8 & 53.5  \\
& \textbf{VPN++ } & $\circ$ & \checkmark &\checkmark  & \textbf{66.8} & \textbf{53.6}  \\
\hline
{\multirow{4}{*}{\rotatebox{90}{\small{New Poses}}}} &2s-AGCN~\cite{2sagcn2019cvpr}&\checkmark& $\times$ & $\times$ & 57.1 & 49.7\\
&VPN~\cite{VPN}  & \checkmark & \checkmark &\checkmark  & 65.2 & 54.1  \\
& \textbf{VPN++} & $\circ$ & \checkmark &\checkmark  & 69.0 & 54.9  \\
 & \textbf{VPN++ + 3D Poses} & \checkmark & \checkmark &\checkmark  & \textbf{71.0} & \textbf{58.1}  \\
\hline
\end{tabular}}
\label{smarthome_accuracy}
}
\end{table}

\begin{table}[t]\tabcolsep=1.4pt
\centering
\caption{Results on NTU-60 dataset with cross-subject (CS) and cross-view (CV) settings (accuracies in \%).}{%
\scalebox{1}{
\begin{tabular}{llccccc}
\hline
&Methods & Pose & RGB & Att & CS & CV  \\
\hline\hline
&2s-AGCN~\cite{2sagcn2019cvpr}& \checkmark & $\times$ & $\times$ & 88.5 & 95.1 \\
& DGNN~\cite{directed_graph} & \checkmark & $\times$ & $\times$ & 89.9 & 96.1 \\
&  MS-G3D Net~\cite{msg3d} & \checkmark & $\times$ & $\times$ & 91.5 & 96.2 \\
\hline
&PEM~\cite{pem} & \checkmark  & \checkmark  & $\times$  & 91.7 & 95.2  \\
\hline
{\multirow{5}{*}{\rotatebox{90}{I3D}}}&Separable STA~\cite{STA_iccv} & \checkmark & \checkmark &\checkmark  & 92.2 & 94.6  \\
&P-I3D~\cite{spatial-i3d} & \checkmark & \checkmark & \checkmark & 93.0 & 95.4 \\
&VPN~\cite{VPN} & \checkmark & \checkmark &\checkmark  & 93.5 & 96.2  \\
& \textbf{VPN++ } & $\circ$ & \checkmark &\checkmark  & 91.9 & 94.9  \\
&  \textbf{VPN++ + 3D Poses} & \checkmark & \checkmark &\checkmark  & \textbf{94.9} & \textbf{98.1}  \\
\hline
{\multirow{4}{*}{\rotatebox{90}{RNX3D}}}&VPN (RNX3D101)~\cite{VPN} & \checkmark & \checkmark &\checkmark  & 95.5 & 98.0  \\
&\small{RNX3D101+MS-AAGCN~\cite{msaagcn}} & \checkmark  & \checkmark  & $\times$  & 96.1 & 99.0  \\
&\textbf{VPN++ } & $\circ$ & \checkmark &\checkmark  & 93.5 & 96.1  \\
&  \textbf{VPN++ + 3D Poses} & \checkmark & \checkmark &\checkmark  & \textbf{96.6} & \textbf{99.1}  \\
\hline
\end{tabular}}
\label{ntu60_accuracy}
}
\end{table}
\begin{table}[t]\tabcolsep=1.8pt
  \centering
\caption{Results on NTU-120 dataset with cross-subject ($CS_1$) and cross-setup ($CS_2$) settings (accuracies in \%); Att indicates attention mechanism.}{%
\begin{tabular}{lccccc}
\hline
Methods & Pose & RGB & Att & $CS_1$ & $CS_2$  \\
\hline\hline
2s-Att LSTM~\cite{global-context} &\checkmark  & $\times$  &\checkmark  & 61.2 & 63.3  \\
Multi-Task CNN~\cite{clip_representation} &\checkmark  & $\times$  & $\times$  & 62.2 & 61.8  \\
PEM~\cite{pem} &\checkmark  & $\times$  &\checkmark  & 64.6 & 66.9  \\
2s-AGCN~\cite{2sagcn2019cvpr} &\checkmark  & $\times$  &\checkmark  & 82.9 & 84.9  \\
 MS-G3D Net~\cite{msg3d} & \checkmark & $\times$ & $\times$ & 86.9 & 88.4 \\
\hline
Two-streams~\cite{twostream} & $\times$ &\checkmark & $\times$ & 58.5 & 54.8 \\
I3D$^*$~\cite{i3d} & $\times$ & \checkmark & $\times$ & 77.0 & 80.1 \\
\hline
\small{Two-streams + ST-LSTM}~\cite{ntu120} & \checkmark & \checkmark & $\times$ & 61.2 & 63.1 \\
Separable STA$^*$~\cite{STA_iccv} & \checkmark & \checkmark &\checkmark  & 83.8 & 82.5  \\
VPN~\cite{VPN} & \checkmark & \checkmark &\checkmark  & 86.3 & 87.8  \\
\hline
 \textbf{VPN++} & $\circ$ & \checkmark &\checkmark  & 86.7 & 89.3  \\
  \textbf{VPN++ + 3D Poses} & \checkmark & \checkmark &\checkmark  & \textbf{90.7} & \textbf{92.5}  \\
\hline
\end{tabular}
\label{ntu120_accuracy}
}
\end{table}
\begin{table}[t]
  \centering
\caption{Results on N-UCLA dataset with cross-view  $V^3_{1,2}$ settings (accuracies in \%); $\overline{Pose}$ indicate its usage only in the training phase.}{%
\begin{tabular}{lccc}
\hline
Methods & Data & Att & $V^3_{1,2}$  \\
\hline\hline
HPM+TM~\cite{hpm} & Depth & $\times$ & 91.9 \\
\small{Ensemble TS-LSTM~\cite{tslstm}} & Pose & $\times$ & 89.2 \\
SGN~\cite{zhang2020semantics} & Pose & $\times$ & 92.5\\
NKTM~\cite{nktm} & RGB & $\times$ & 85.6 \\
I3D$^*$~\cite{i3d} & RGB & $\times$ & 86.0 \\
Glimpse Cloud~\cite{glimpse} & RGB+ $\overline{Pose}$ & \checkmark & 90.1 \\
Separable STA~\cite{STA_iccv} & RGB+Pose& \checkmark & 92.4  \\
P-I3D~\cite{spatial-i3d} & RGB+Pose & \checkmark & 93.1  \\
Global Model~\cite{wacv_temporal} &  RGB+Pose & \checkmark & \textbf{93.5} \\
VPN~\cite{VPN} & RGB+Pose& \checkmark & \textbf{93.5}  \\
\hline
 \textbf{VPN++ } & RGB+ $\overline{Pose}$ & \checkmark & 91.9 \\
  \textbf{VPN++ + 3D Poses} & RGB+Pose & \checkmark & \textbf{93.5} \\
\hline
\end{tabular}
\label{ucla_accuracy}
}
\end{table}

\section{Comparison to the the state-of-the-art}
We compare VPN++ to the State-of-the-Art (SoA) on Smarthome, NTU-60, NTU-120, and N-UCLA in Tables~\ref{smarthome_accuracy}, \ref{ntu60_accuracy}, \ref{ntu120_accuracy}, and~\ref{ucla_accuracy}. 

For smarthome dataset, we present the SoA categorized into RGB and RGB+Pose based methods in Table~\ref{smarthome_accuracy}. 
We provide the evaluation results on old Poses and new Poses (referred to as low \& medium levels of Pose quality in Table~\ref{pose_quality_2}). VPN++ outperforms all the SoA methods by up to 9.8\%  \& 5.8\% relatively with old and new Poses respectively. This significant improvement on this dataset can be explained by the video-pose embedding infused through our two levels of distillation for combining RGB \& Poses, which in turn handles the challenge of low camera framing~\cite{STA_iccv}. As discussed earlier, often low quality Poses are obtained in real-world scenarios with occlusions and low subject resolution. Thanks to the distillation mechanisms, it encourages the classification model to selectively infuse the relevant Pose information into the RGB stream.
by providing a discriminative video-pose embedding. 
For NTU-60 dataset, VPN++ achieves accuracy close to the methods requiring Poses at test time whereas for NTU-120, VPN++ outperforms the later. We observe that the skeleton based action recognition methods perform better compared to the RGB based methods on NTU dataset. But this is due to the high quality of Poses (with no occlusion) which makes the dataset apt for Pose only methods. On the contrary, in real-world dataset like Smarthome (see Table~\ref{smarthome_accuracy}), the Pose only methods substantially under-perform compared to the RGB based methods. Another limitation of Pose only methods includes their lack of appearance encoding. However, VPN++ when combined with 3D Poses outperforms SoA on both NTU datasets. We confirm the robustness of VPN++ by evaluating it with 3D ResNext-101~\cite{can_spatio-temporal} as a video backbone on NTU-60.
Similar observations can also be done on N-UCLA dataset in Table~\ref{ucla_accuracy} hinting that VPN++ generalizes over small scale datasets too.
\begin{table}[h!]
\caption{Choice of models to the practitioners. $\uparrow$ indicates High and $\downarrow$ indicates Low with (A) inference time, (B) quality of poses, (C) model size, and (D) amount of training data. Accuracy is provided for different Datasets.}
\scalebox{0.78}{
\begin{tabular}{c|c|c|c|c|c}
\hline
Settings  & Model Choice & Dataset & Inference & \# & Acc.\\
Criterion &              &         & time     & \small{Param.} & (in \%)\\\hline
Baseline (RGB)      &     I3D~\cite{i3d}  & SH  & 0.3s  & 12M &  53.4   \\
Baseline (Pose)      &     2s-AGCN~\cite{2sagcn2019cvpr}  & SH  & 64s  & 3.5M &  57.1   \\
SoA       &     VPN  & SH  & 65s  & 24M &  65.2   \\
A ($\downarrow$) or B ($\downarrow$) & VPN++ & SH  & 0.4s & 14M & 69.0  \\
B ($\uparrow$) or C ($\downarrow$)  & VPN++ + Poses & SH & 64.4s & 17.5M & 71.0  \\
 B ($\downarrow$) or C ($\downarrow$) & VPN-F & SH & 0.3s & 12M & 67.1  \\ 
A ($\downarrow$), C ($\downarrow$), D ($\uparrow$)  & VPN++ & NTU-120 & 0.35s & 14M & 86.7  \\ 
A ($\downarrow$), C ($\downarrow$), D ($\downarrow$)  & VPN-F & NTU-60 & 0.28s & 12M & 90.8  \\
\hline
\end{tabular}}
\vspace{-0.1in}
\label{choices}
\end{table}
\section{Discussion}
In this paper, we extend our previous framework Video-Pose network (VPN) to explore new mechanisms for combining video and Poses in order to classify action.
Consequently, we have proposed two levels of distillation that can be adapted to different real-world application settings for recognizing actions. 
We summarize in table~\ref{choices}, the appropriate choice of distillation model or fusion mechanisms that could be exploited based on the requirements of a practitioner. Along with providing the appropriate choice of models, we also present the inference time, number of parameters of the resultant model, and action classification accuracy on relevant datasets.
The choice of a model is based on factors (i.e. application requirements or network settings) concerning its applicability like (A) inference time, (B) quality of poses, (C) model size, and (D) amount of training data. From this experimental analysis, we conclude that our variants of distillation model (i.e., VPN++ and VPN-F) are useful when the end-user wants real-time predictions (e.g., low inference time), whereas the late fusion of VPN++ and Poses is preferred for offline action recognition. We notice that VPN-F is an effective model if further speed-up is required compared to VPN++ under the constraints of bad quality of  Poses or less available training data.
Interestingly, these lighter models are more accurate 
than models with similar training modalities~\cite{VPN, glimpse, spatial-i3d, STA_iccv, wacv_temporal}.
\begin{table}[t]
  \centering
\caption{Effectiveness of Video-Flow Network++ (VFN++) representation using our SCD loss.}{%
\scalebox{0.95}{
\begin{tabular}{lccc}
\hline
Stream & SH   & NTU-60  & NTU-60 \\
      & (CS)  & (CS)    & (CV)    \\
\hline
RGB &  53.4 & 85.5 & 87.3  \\
OF & 51.8 & 85.7 & 92.8 \\
RGB + OF & 57.3 & 87.1 & 93.6 \\
\hline
MARS + RGB~\cite{mars} & 58.1 & 88.2 & 92.9 \\
VFN++ &  59.0 & 90.1 & 93.4   \\
VFN++ + OF &  \textbf{66.4} & \textbf{94.6} & \textbf{97.2}   \\
\hline
\end{tabular}
}
\label{flow}
}
\end{table}
\begin{table}[t]
  \centering
\caption{Combination of RGB, 3D Poses and Flow modalities into a single model. Here VPFN++ is VPN++ + VFN++.}{%
\scalebox{1}{
\begin{tabular}{lccc}
\hline
Fusion & SH   & NTU-60  & NTU-60 \\
      & (CS)  & (CS)    & (CV)    \\
\hline
RGB + OF + 3D Poses &  64.4 & 90.2 & 95.9  \\
VPN++ & 69.0   &  91.9  &  94.9 \\
\hline
VPFN++ & 69.7 & 92.1 & 95.5 \\
VPFN++ + 3D Poses & 71.7 & 95.1 & 98.2 \\
VPFN++ + 3D Poses + OF & \textbf{72.9} & \textbf{96.7} & \textbf{99.1}  \\
\hline
\end{tabular}
}
\label{combine}
}
\end{table}
\section{Scope beyond RGB and Poses}
In this section, we go beyond utilizing video-pose embedding by combining video with other modalities like Optical Flow through distillation. 
To this end, we investigate the applicability of using the distillation mechanisms involved in VPN++ for combining RGB and Optical Flow (OF). In our experiments, following the attempts towards distillation at feature space level~\cite{mars}, we use supervised contrastive distillation loss (SCD) between the features of RGB and OF streams. We dub this new Flow Augmented RGB stream as VFN++. Note that the Flow backbone for this experiment is an I3D flow stream. Experimentally, we find that attention level distillation is not effective while using optical flow as a Teacher. This might be due to high dimensionality of the flow features that hampers the attention network to learn relevant attention weights. Moreover, the flow features are not view-adaptive and do not consider the human anatomy while learning the attention weights. We present a comparative study with VFN++ in Table~\ref{flow}. We observe that VFN++ outperforms MARS+RGB due to the supervised contrastive learning mechanism. On availability of OF at inference time, the performance shoots up significantly. However this accuracy is lower than the one obtained with 3D Poses, substantiating that 3D Poses are superior than OF for ADL with subtle actions. In Table~\ref{combine}, we take a step forward towards combining VPN++ and VFN++. We call this resultant model Video-Pose-Flow Network++ (VPFN++). The combination is performed by the late fusion of VPN++ and VFN++ prediction scores. The minor performance improvement (+0.7\% for Smarthome \& +0.4\% for NTU-60) in VPFN++ compared to our distillation model (VPN++) is attributed to OF distillation. So, for ADL, OF does not contribute much when 3D Poses are already well infused in RGB. 
With availability of Poses and OF at test time, VPFN++ + 3D Poses + OF supersedes the SoA models. Thus, our proposed framework could be extended for combining privileged modalities which is a possible  perspective of this work. However, it is to be introspected that 
the appropriate distillation mechanism may depend on the given modalities.   

\section{Conclusion}
\vspace{-0.05in}
In this paper, we have extended our proposed video-pose embedding for video understanding and presented a different perspective for combining RGB and 3D Poses through knowledge distillation. 
In an attempt to rethink combining RGB and Poses via feature fusion and attention mechanism, we propose two levels of distillation by infusing Poses at training time - feature-level and attention-level. Consequently, VPN++ does not rely anymore on the availability of 3D poses at inference time resulting in high speed up and high resiliency to noisy Poses. In addition to this, VPN++ learns a discriminative representation for classifying ADL. 
We show that VPN++ when combined with 3D Poses, if available, outperforms the state-of-the-art methods on 4 ADL datasets. Then, we study different strategies of combining modalities for video understanding which could be exploited by the community based on their needs.

\noindent Preliminary results show also that VPN++ can be extended to optical flow.
Future work will explore towards an end-to-end framework, infusing several modalities simultaneously into a RGB stream.

\section*{Acknowledgement}
We are grateful to INRIA Sophia Antipolis - Mediterranean "NEF" computation cluster for providing resources and support.

\bibliographystyle{IEEEtran}
\bibliography{egbib}



%





\end{document}